\documentclass[journal]{IEEEtran}
\usepackage{times}
\usepackage{soul}
\usepackage{url}
\usepackage[hidelinks]{hyperref}
\usepackage[utf8]{inputenc}
\usepackage[small]{caption}
\usepackage{graphicx}
\usepackage{amsmath}
\usepackage{amsthm}
\usepackage{booktabs}
\usepackage{algorithm}
\usepackage{algorithmic}
\usepackage{amssymb}
\usepackage{times}
\usepackage{epsfig}
\usepackage{booktabs}
\usepackage{longtable}
\usepackage{subfigure}
\usepackage{setspace}
\usepackage{bm}
\usepackage{cite}
\usepackage{multirow}
\usepackage{balance}

\ifCLASSINFOpdf
\else
\fi
\begin{document}
\title{Not All Instances Contribute Equally: Instance-adaptive Class Representation Learning for Few-Shot Visual Recognition}

\author{Mengya Han,
        Yibing Zhan,
        Yong Luo,
        Bo Du,
        Han Hu,
        Yonggang~Wen~\IEEEmembership{Fellow,~IEEE},
        and~Dacheng~Tao~\IEEEmembership{Fellow,~IEEE} %~\IEEEmembership{Fellow,~IEEE}% <-this % stops a space
%\thanks{M. Han, Y. Luo and B. Du are with the National Engineering Research Center for Multimedia Software, School of Computer Science, Institute of Artificial Intelligence and Hubei Key Laboratory of Multimedia and Network Communication Engineering, Wuhan University, Wuhan 430072, China (e-mail: myhan1996@whu.edu.cn; yluo180@gmail.com; dubo@whu.edu.cn). (Corresponding author: Yong Luo.)}% <-this % stops a space
\thanks{M. Han, Y. Luo and B. Du are with the National Engineering Research Center for Multimedia Software, School of Computer Science, Wuhan University, and Hubei Luojia Laboratory, Wuhan, China (e-mail: myhan1996@whu.edu.cn; yluo180@gmail.com; dubo@whu.edu.cn). This work was done when Mengya Han was visiting JD Explore Academy as a research intern. Corresponding author: Yong Luo.}% <-this % stops a space
%\thanks{Corresponding author: Yong Luo.}
\thanks{Y. Zhan and D. Tao are with JD.com (e-mail: zhanyibing@jd.com, dacheng.tao@gmail.com).}% <-this % stops a space
\thanks{H. Hu is with the School of Information and Electronics, Beijing Institute of Technology, Beijing 100081, China (e-mail: hhu@bit.edu.cn).}
\thanks{Y. Wen is with the School of Computer Science and Engineering, Nanyang Technological University, Singapore 639798 (e-mail: ygwen@ntu.edu.sg).}
%\thanks{Corresponding authors: Yong Luo; Bo Du.}
}


\markboth{IEEE Transactions on Neural Networks and Learning Systems}%
%\markboth{$>$ \normalsize{A S}\footnotesize{ubmission to} \normalsize{IEEE T}\footnotesize{ransactions on} \normalsize{N}\footnotesize{eural} \normalsize{N}\footnotesize{etworks and} \normalsize{L}\footnotesize{earning} \normalsize{S}\footnotesize{ystems} $<$}%
{Shell \MakeLowercase{\textit{et al.}}: }



\maketitle

\begin{abstract}

Few-shot visual recognition refers to recognize novel visual concepts from a few labeled instances. Many few-shot visual recognition methods adopt the metric-based meta-learning paradigm by comparing the query representation with class representations to predict the category of query instance. However, current metric-based methods generally treat all instances equally and consequently often obtain biased class representation, considering not all instances are equally significant when summarizing the instance-level representations for the class-level representation. For example, some instances may contain unrepresentative information, such as too much background and information of unrelated concepts, which skew the results. To address the above issues, we propose a novel metric-based meta-learning framework termed instance-adaptive class representation learning network (ICRL-Net) for few-shot visual recognition. Specifically, we develop an adaptive instance revaluing network with the capability to address the biased representation issue when generating the class representation, by learning and assigning adaptive weights for different instances according to their relative significance in the support set of corresponding class. Additionally, we design an improved bilinear instance representation and incorporate two novel structural losses, i.e., intra-class instance clustering loss and inter-class representation distinguishing loss, to further regulate the instance revaluation process and refine the class representation. We conduct extensive experiments on four commonly adopted few-shot benchmarks: miniImageNet, tieredImageNet, CIFAR-FS, and FC100 datasets. The experimental results compared with the state-of-the-art approaches demonstrate the superiority of our ICRL-Net.
\end{abstract}


% \fi
\IEEEpeerreviewmaketitle
\section{Introduction}
\IEEEPARstart{D}{eep} learning models have achieved the state-of-the-art performance on various visual tasks, including image classification~\cite{he2016deep,huang2017densely,hu2018squeeze}, object detection~\cite{ren2015faster,redmon2016you}, and segmentation~\cite{he2017mask}.
However, most deep models have numerous parameters and require large amounts of labeled data for training. In most cases, obtaining abundant labeled data is time-consuming and laborious.
The human annotation cost and data scarcity in some classes (\textit{e.g.}, rare species) significantly limit the applicability of current vision systems to learn new visual concepts efficiently.
In contrast, learning from extremely few labeled instances is an important ability for humans.
%For example, children can easily form the concept of ``giraffe" by only taking a glance from a picture or hearing its description as ``giraffe looks like a deer with a long neck."
It is thus of great interest to develop machine learning algorithms that recognize new visual categories from only a limited amount of labeled instances for each novel category. The problem of learning to recognize unseen classes from limited instances, known as few-shot learning (FSL)~\cite{fe2003bayesian,fei2006one}, has attracted increasing attention recently.

\begin{figure}[!t]
    \center
    \includegraphics[width=1.0\linewidth]{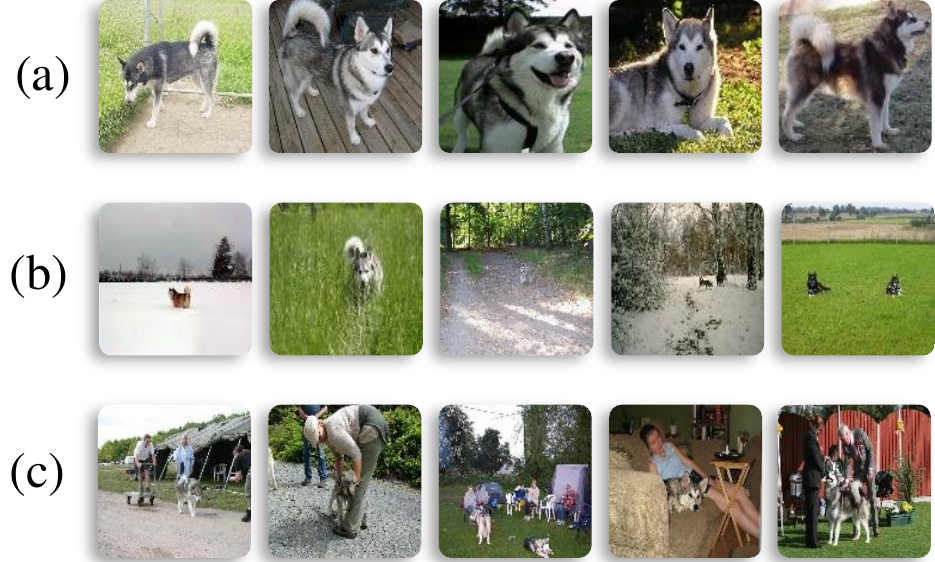}
   \caption{Three types of support instances of one class: (a) shows representative support instances; (b) and (c) provide support instances that contain too much background and information of unrelated concepts, respectively. Intuitively, the more informative instances, such as the instances shown in row (a), will be more useful when obtaining class representations. Whereas, the instances in (b) and (c) should be revalued (often to be less important) in obtaining optimal class representation.}
   \label{fig:intro}
\end{figure}

Few-shot visual recognition~\cite{9159937,9072304,8948295,gidaris2018dynamic,pmlr-v119-ziko20a,lee2020self,zhang2020uncertainty,Luo_2021_WACV,li2021multi}, as a specific FSL problem, attempts to learn a classifier with good generalization ability for novel visual concepts, each of which contains only a few labeled instances (support instances). To address this problem, a variety of solutions have been proposed. One mainstream solution is meta-learning~\cite{vinyals2016matching,ravi2016optimization,lee2019meta,9174820}, where a series of independent few-shot tasks are utilized to learn a general model during training, and then the general model is applied to unseen target tasks in the testing phase. In this way, the learned meta-models solve a new task by utilizing the knowledge acquired from many similar tasks. Generally, meta-learning methods for few-shot visual recognition can be classified into two categories: optimization-based and metric-based. The former~\cite{finn2017model,rusu2018meta,elsken2020meta} automatically learns a set of model parameters, such as general initialization conditions, learning rates, and parameter updating strategies. The latter~\cite{2020Adaptive,tian2020rethinking,luo2021few} aims to learn a discriminative embedding space, in which the representations of different instances and classes can be easily distinguished. Compared with optimization-based approaches, metric-based methods are more efficient and widely applicable, and thus we focus on metric-based ones in this paper.

In metric-based meta-learning methods for few-shot visual recognition, the instance representations are first extracted (such as by using neural networks). Then the class representations are obtained by utilizing the instances in the corresponding support set. Afterward, the query instances are classified by comparing the query representations with the obtained class representations. However, existing metric-based methods~\cite{snell2017prototypical,sung2018learning,2020A} often regard that all instances within the same support set are equally significant when obtaining the class-level representation. They ignore the information diversity that exists between the support instances and often obtain biased class representations, and hence sub-optimal performance can be achieved. For example, Fig.~\ref{fig:intro} provides three types of support instances of one class: (a) shows representative support instances, and (b) and (c) provide support instances that contain too much background and information of unrelated concepts (such as the person category) respectively. Intuitively, the more informative instances, such as the instances shown in Fig.~\ref{fig:intro} (a), will be more useful and should be revalued to be more important when obtaining the optimal class representation, while the instances in (b) and (c) should be revalued to contribute less.

This motivates our instance-adaptive class representation learning network (ICRL-Net), which adaptively revalues the importance of different instances in the same class when obtaining the class-level representation. The main contributions of this paper are:

\begin{enumerate}
 \item We propose a novel metric-based meta-learning method, ICRL-Net, for few-shot visual recognition. Specifically, ICRL-Net contains an attentional bilinear feature extraction module that generates efficient instance-level representations, which can reduce the computational cost and improve the subsequent instance revaluation.
 %, an adaptive instance revaluing network that revalues the significance of different instances for better class-level representation, and a classifier based on a designed joint loss.
 %\item We propose a novel metric-based meta-learning method, ICRL-Net, for few-shot visual recognition. Specifically, ICRL-Net contains an attentional bilinear feature extraction module that generates efficient instance-level representations, an adaptive instance revaluing network that revalues the significance of different instances for better class-level representation, and a classifier based on a designed joint loss.
 %that performs the final classification by comparing the query instance representations with the learned class representations.

 \item We develop an adaptive instance revaluing network to alleviate the issue of bias in class representation. Specifically, the adaptive instance revaluing network assigns different weights (or values) as the relative significance of instances by considering all support instances in the same classes and obtains the class representation by averaging the revalued instance representations.
 %(i.e., the product of instance value and instance representation).

 \item We design a joint loss function to improve the instance adaptive revaluation process and refine the discriminative representation space of metric-based meta-learning. The designed joint loss function contains three components: a commonly adopted classification loss for few-shot visual recognition and two newly designed structural losses for robust representation: intra-class instance clustering loss and inter-class representation distinguishing loss.
\end{enumerate}

We conducted extensive experiments on four popular few-shot benchmarks: miniImageNet~\cite{vinyals2016matching}, tieredImageNet~\cite{ren2018meta}, CIFAR-FS~\cite{bertinetto2018meta}, and FC100~\cite{oreshkin2018tadam} datasets.
The experimental results indicate that our ICRL-Net outperforms the state-of-the-art approaches. In specific, we achieved a significant 1.5\% relative improvement compared with the most competitive counterpart.
We further visualize the learned attention parts of ICRL-Net, and the results show that our ICRL-Net can appropriately identify the relative significance between instances in the same support set.

The rest of this paper is organized as follows. Section~\ref{sec:Related_Work} briefly introduces the related works. The details of ICRL-Net are depicted in Section~\ref{sec:ICRL-Net}. Section~\ref{sec:Experiments} provides the experimental analysis, and we conclude this paper in Section~\ref{sec:Conclusion}.

\section{Related Work}\label{sec:Related_Work}

This section briefly introduces the few-shot learning and the attention mechanism.

%we will first review the mainstream methods of few-shot learning, which can be roughly grouped into metric-based methods, optimization-based methods. Then, we briefly introduce attention mechanisms and explore how to use attention mechanisms for few-shot learning.
%In this section, we review the literature on few-shot learning for visual recognition tasks.
\subsection{Few-shot learning}
Few-shot learning (FSL) is a emerging research topic that aims to learn a model from a set of data (base classes) and adapt the model to a disjoint set (new classes) with limited training data~\cite{gidaris2019generating,9056801}. Up to now, few-shot visual recognition~\cite{guo2020attentive,chen2021pareto,wang2021mtunet,li2020adversarial}, which aims to recognize novel visual categories from limited labeled instances, has received great attentions in FSL. Earlier work on few-shot learning tended to involve generative models with complex iterative inference strategies~\cite{fei2006one}. Most of the recent FSL approaches follow the meta-learning paradigm, which is usually performed by training a meta-learner that learns the transferable knowledge from similar tasks and then generalizes to new tasks. Under this paradigm, various meta-learning methods for few-shot learning are developed, and can be roughly classified into two categories: optimization-based methods and metric-based methods.

Optimization-based methods~\cite{finn2017model,ravi2016optimization,li2017meta,rusu2018meta,sun2019meta,mishra2017simple} aim to find a single set of model parameters that can be adapted with a few steps of gradient descent to target tasks. For example, the well-known MAML approach~\cite{finn2017model} meta-learns a good initial condition (a set of neural network weights), which enables the model to quickly adapt to new tasks. The few-shot optimization approach Meta-LSTM~\cite{ravi2016optimization} go further to not only learn a good initial condition but also an LSTM-based meta-learner that is utilized to learn appropriate parameter updating rules. Meta-SGD~\cite{li2017meta} further improves the meta-learning ability by learning the parameter initialization, gradient update direction, and learning rate within a single step. Although effective sometimes, using only a few instances to compute gradients in a high-dimensional parameter space could make generalization difficult. This issue is addressed by latent embedding optimization (LEO)~\cite{rusu2018meta}, where a low-dimensional latent embedding of model parameters is learned and optimization-based meta-learning is performed in the embedding space. It turns out that the performance of meta-learning in low-dimensional parameter space is much better than that of meta-learning in high-dimensional space. Meta transfer learning (MTL)~\cite{sun2019meta} leverages the idea of transferring pre-trained weights and learns to effectively transfer large-scale pre-trained deep neural network weights for solving few-shot tasks. Overall, the approaches mentioned above still need to be fine-tuned on the target tasks. In contrast, metric-based methods solve target tasks without any model updates, thus avoid gradient computation during testing.

Metric-based methods~\cite{koch2015siamese, vinyals2016matching, snell2017prototypical, 2017Low, hariharan2017low, sung2018learning, oreshkin2018tadam, li2019revisiting} aim to learn a discriminative representation space, in which the distances between samples should be small in the same class, and large otherwise.
%and the distances between different classes should be small and large, respectively.
The classification is performed in the space by simply finding the nearest neighbor of the query. For example, Koch et al.~\cite{koch2015siamese} calculate correlations between input instances via supervised metric based on siamese neural networks, and then predict the most relevant classes for query instances. Vinyals et al.~\cite{vinyals2016matching} design a matching network that introduces an episodic training strategy for few-shot learning, and trains a neural network to embed examples. Additionally, an attention mechanism was used over the learned representations of the support set to predict the labels of the query set, which can be interpreted as a weighted nearest neighbor classifier. The popular prototypical network~\cite{snell2017prototypical} is built upon~\cite{vinyals2016matching}, which takes a class's prototype to be the mean of its support set in the representation space. Then, it calculates the distances between the class prototypes and the query representation to predict the category for a query instance. Inspired by semi-supervised clustering, Ren et al.~\cite{ren2018meta} propose an extension of the prototypical network, which uses massive unlabeled instances to generate refined prototypes. These approaches focus on learning representations for data such that they can be recognized with a fixed metric~\cite{vinyals2016matching,snell2017prototypical} or linear~\cite{snell2017prototypical,koch2015siamese} classifier. In the relation network~\cite{sung2018learning}, a deep distance metric is learned for comparing the relation between the query instances and the support instances. The metric is learnable and equivalent to a non-linear classifier. Cross attention network~\cite{hou2019cross} is developed based on the relation network, where an attention module is designed to highlight the correct region of interest in query instance to help classification. Meta-Baseline~\cite{2020A} further improves the ability of metric-based methods by pre-training a classifier on all base classes and meta-learning on a nearest-centroid based few-shot classification algorithm. FEAT~\cite{ye2020few} takes advantage of the set-to-set function to generate task-adaptive feature representations. MCT~\cite{kye2020meta} meta-learn the confidence for each query sample and then update class prototypes for each transduction step by using all the query examples with meta-learned scores. ReMP~\cite{zhao2021remp} proposes to refine the prototypes by considering the similarity information of the support set and query set to rectify the metric space, which aims to reduce the metric inconsistently between the training and testing phase.
ICI~\cite{wang2020instance,wang2021trust} assumes that not every unlabeled query instance is equal important and aims to exploit the unlabeled instances to augment the training set. It measures the credibility of pseudo-labeled examples and selects the most trustworthy pseudo-labeled samples according to their credibility as augmented labeled instances.
However, most of the existing metric-based methods generally ignore the negative influence of support instances that contain too much interference information and often obtained biased class representations~\cite{snell2017prototypical,2020A}. To address this issue, we propose a metric-based meta-learning method, ICRL-Net, which refines the class representation by revaluing the significance of different support instances based on the attention mechanism.

\subsection{Attention Mechanism}
The attention mechanism can focus on the discriminative area adaptively and has been widely exploited for various tasks~\cite{hu2018squeeze, Wang_2018_CVPR,hou2019cross}. For example, Hu et al.~\cite{hu2018squeeze} propose the squeeze-and-excitation network (SENet) to weight each channel of the feature map. Woo et al.~\cite{woo2018cbam} propose convolutional block attention module (CBAM), which employs hybrid spatial and channel features for attention design. These attention blocks either focus on the channel encoding or spatial context connection. In addition, there are other forms of attention mechanisms, such as graph attention~\cite{velivckovic2017graph} and self-attention~\cite{vaswani2017attention, han2021transformer}.
In few-shot learning, there are many works~\cite{Wang_2017_CVPR,vinyals2016matching,hou2019cross,ye2020few,xu2021attentional} also adopts the attention mechanism and achieved excellent performance. For example, in the matching network~\cite{vinyals2016matching}, attention mechanism is utilized together with the softmax function to fully specify the prediction of the meta-learner classifier. Similar to the matching network, the cross attention network~\cite{hou2019cross} models the semantic dependency between support instances and query instances. Hence, the relevant regions on the query instances are adaptively localized so that the discrimination ability of embedding features can be improved. In general, matching network focuses on embeddings while cross attention network manipulates feature maps. FEAT~\cite{ye2020few} utilizes the self-attention~\cite{vaswani2017attention} to generate task-adaptive feature representations. In contrast to these works, our developed AIRN mainly focus on learning robust class prototypes by identifying important support instances and suppressing irrelevant information.
%attempts to model the relationships between support instances within the same class and focus on the more important instances adaptively.

\begin{figure*}
   \centering
   \includegraphics[width=1.0\textwidth]{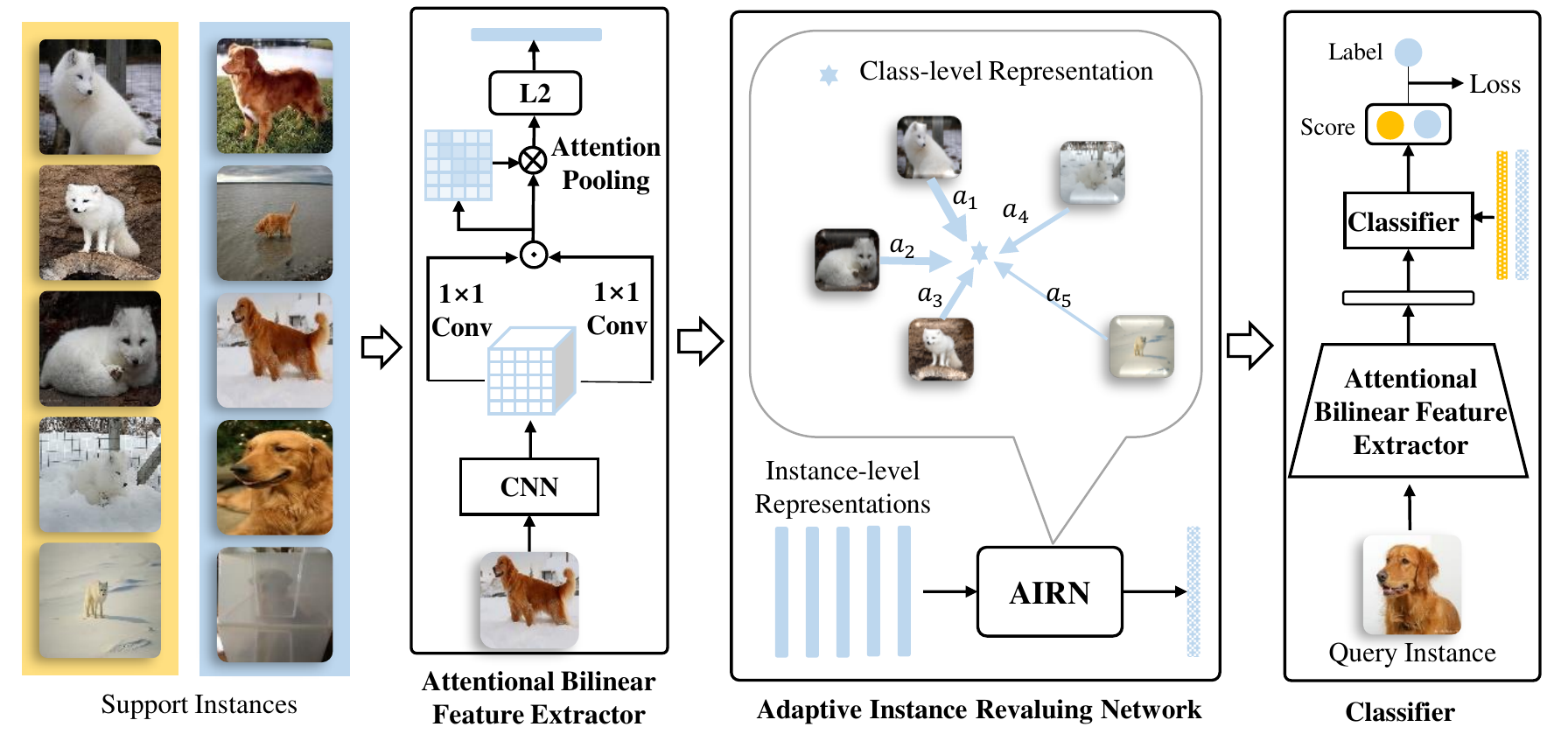}
   \caption{Overview of the proposed ICRL-Net. ICRL-Net consists of an attentional bilinear feature extractor, an adaptive instance revaluing network, and a classifier. Specifically, ICRL-Net first extracts instance-level representations through the attentional bilinear feature extractor. Note that both support and query instances share the same feature extractor. Then, ICRL-Net uses the adaptive instance revaluing network to assign different values of support instances' relative significance and obtain the class-level representations from few-shot support instances. Finally, the query instances are classified by comparing the query instances' representations with the learned class representations of support instances.
}
\label{fig:archtecture}
\end{figure*}

%\section{The Proposed ICRL-Net}\label{sec:4}
\section{Instance-adaptive Class Representation Learning} \label{sec:ICRL-Net}
This section depicts our Instance-adaptive Class Representation Learning Network (ICRL-Net). %Specifically, we first introduce the problem formulation and then explain the details of the ICRL-Net.

% This section presents the Instance-adaptive Class Representation Learning Network (ICRL-Net). Specifically, we first introduce the problem formulation of few-shot learning, and then explain the details of the ICRL-Net, sequtially including the framework, the attentional bilinear feature extractor, the adaptive instance revaluing network, and the classifier.

\subsection{Problem Formulation}
We first introduce the problem formulation of FSL. Let $\mathcal{S}$ denote the support set, which contains $N$ classes and $K$ labeled instances per class. Then, given a query set $\mathcal{Q}$, FSL aims to determine the class of each unlabeled instance in $ \mathcal{Q}$ based on the support set $\mathcal{S}$. The setting mentioned above is also called the $N$-way $K$-shot classification task.

In FSL, the episodic training strategy~\cite{vinyals2016matching} is often utilized to obtain a well-trained model by sampling few instances for every episode (an individual $N$-way $K$-shot task) under the meta-learning paradigm. We follow the episodic training strategy, which efficiently learns transferable knowledge from a relatively large labeled dataset $\mathcal{D}_{train}$ that contains a set of classes $\mathcal{C}_{train}$. Then, the trained model is applied to a novel testing dataset $\mathcal{D}_{test}$ that contains a set of classes $\mathcal{C}_{test}$. There are only a few labeled instances provided for each category in $\mathcal{C}_{test}$, and $\mathcal{C}_{train} \bigcap \mathcal{C}_{test}= \varnothing$. The process of training is conducted on a series of episodes.
In each episode, a small subset of $N$ classes are sampled from $\mathcal{C}_{train}$ to construct an $N$-way $K$-shot task: a support set $ \mathcal{S} = \left \{ \left (x_{n}^{k} ,y_{n}^{k}\right ) |n=1,...,N; k= 1,...,K\right \} $ and a query set $ \mathcal{Q} = \left \{ \left (q_{n}^{m} ,y_{n}^{m}\right ) |n=1,...,N; m= 1,...,M\right \} $, where $N$ is the number of classes in each episode, $K$ is the number of support instances in each class, and $M$ is the number of query instances in each class. In each episode, the model is trained by minimizing the prediction loss of the query set $\mathcal{Q}$.

In the testing, the generalization performances of the learned models are measured on the testing set episodes, each of which consists of a support set $\mathcal{S}$ and a query set $ \mathcal{Q}$, where $\mathcal{S}$ and $\mathcal{Q}$ are sampled from $\mathcal{D}_{test}$ that contains classes distinct from those used in $\mathcal{D}_{train}$. The instance labels in the support set are known, while those in the query set are unknown and used only for evaluation. The predicted label of the query instance is given by taking the class that has the highest classification score.
%The accuracy is therefore obtained by comparing the predicted labels with true labels for each episode.

\subsection{Overview}

Fig.~\ref{fig:archtecture} presents the overview of our ICRL-Net. As shown in Fig.~\ref{fig:archtecture}, ICRL-Net consists of three modules: an attentional bilinear feature extractor to extract instance-level representations, an adaptive instance revaluing network (AIRN) to learn the relative significance of support instances in the same class and accordingly obtains the class-level representations, and a classifier to classify each query instance based on the learned class-level representations. In addition, ICRL-Net is trained based on a designed joint loss, which contains a commonly adopted classification loss and two newly designed structural losses: intra-class instance clustering loss and inter-class representation distinguishing loss. More details of the different components and losses are explained as follows.
%In the remaining subsections, we will sequentially introduce the feature extraction network, the adaptive instance revaluing network, the classifier, and the joint loss.

\subsection{Attentional Bilinear Feature Extractor}

Metric-based methods~\cite{vinyals2016matching,snell2017prototypical,2020A} perform classification by comparing the (support and query) instance representations, and thus it is critical to learn discriminative instance representation.
It is well known that exploiting high-order information~\cite{lin2015bilinear,ionescu2015matrix, kim2016hadamard}, such as by utilizing bilinear pooling, can improve the discriminative capability of feature representations compared with the low-order information~\cite{lin2013network}.
Several few-shot works recently exploited a variety of variants of bilinear pooling to learn the better feature representation. For example, Second-order Similarity Network~\cite{zhang2019power} leverages second-order pooling (i.e., Homogeneous Bilinear Pooling) to learn the second-order statistics for similarity learning. Following SoSN, Saliency-guided Hallucination Network~\cite{zhang2019few}, MsSoSN~\cite{zhang2020few}, and Few-shot Localizer~\cite{wertheimer2019few} also employ second-order statistics to improve the accuracy which demonstrates the usefulness of second-order pooling in few-shot learning. The recent work~\cite{koniusz2021power} also gives a detailed theoretical analysis to demonstrate that the element-wise product of feature pairs can learn the correlations (higher-order statistics). Therefore, the element-wise product of the outputs of two 1$\times$1 convolutions can capture the higher-order statistics (correlations) for learning better feature representation.
%Bilinear pooling has been commonly exploited in various visual applications~\cite{Gao_2016_CVPR,Kong_2017_CVPR,Yu_2018_ECCV,Liao_2019_ICCV}. However, the original bilinear pooling uses the outer product of instance features as the final instance representation. This results in very high-dimensional features, which are prone to be over-fitting and often lead to high computational costs.
%Inspired by bilinear pooling, in this paper, we propose an efficient feature extractor, namely, the attentional bilinear feature extractor, for better feature representation and to improve the revaluation process of AIRN.
To learn a better instance-level feature representation and improve the revaluation process of AIRN, we propose an efficient feature extractor, namely, the attentional bilinear feature extractor.
As shown in Fig. \ref{fig:archtecture}, our attentional bilinear feature extractor consists of a convolutional neural network followed by two parallel $1 \times 1$ convolutional layers, an attention pooling, and an L2 normalization.

Specifically, given an input instance $x_{n}^{k}$, its initial feature representation is the output of a convolutional neural network given by $f\left (x_{n}^{k} \right)\in \mathbb{R}^{d\times h\times w}$, where $d$, $h$, and $w$ are the channel, height, and width number of the instance feature, respectively. Then two parallel convolutional layers with kernel size $1$ are applied to each initial representation to generate two feature representations $f_1=f_{w_1}(f(x_{n}^{k}))\in \mathbb{R}^{ d\times h \times w}$ and  $ f_2=f_{w_2}(f(x_{n}^{k})) \in \mathbb{R}^{ d\times h \times w}$, where ${w}_1$ and ${w}_2$ are parameters of two $1 \times 1$ convolutional layers. Afterward, we compute the Hadamard product of two feature representations $f_1$ and $f_2$ to generate an intermediate feature representation $\hat{ f(x_{n}^{k})}=f_1 \odot f_2$, where $\odot$ denotes the element-wise multiplication.
Attention pooling is applied on the intermediate representation $\hat{ f(x_{n}^{k})}$ for dimension reduction, where the spatial-wise weight $A_s \in \mathbb{R}^{h\times w}$ for attention pooling is calculated as:
\begin{equation}
A_s = \delta( f_{w_s}(\hat{f(x_{n}^{k})})),
\end{equation}
where $w_{s}$ denotes the parameter of a $1 \times 1$ convolutional layer and $\delta$ is the sigmoid function.
Formally, the intermediate feature representation can be reshaped as $ \hat{ f(x_{n}^{k})} \in \mathbb{R}^{ d\times hw}$, and the spatial-wise attention weights can be reshaped as $A_s \in \mathbb{R}^{hw \times 1}$.
Therefore, the final instance-level representation $F_{n}^{k} \in \mathbb{R}^{d}$ of instance $x_{n}^{k}$ is given by
\begin{equation}
F_{n}^{k}=L_2(\hat{f(x_{n}^{k})} \cdot A_s).
\end{equation}
where $L_2$-normalization is adopted. Both the query and support instances share the same feature extractor.

\begin{figure}[!t]
    \centering
    \includegraphics[width=0.95\linewidth]{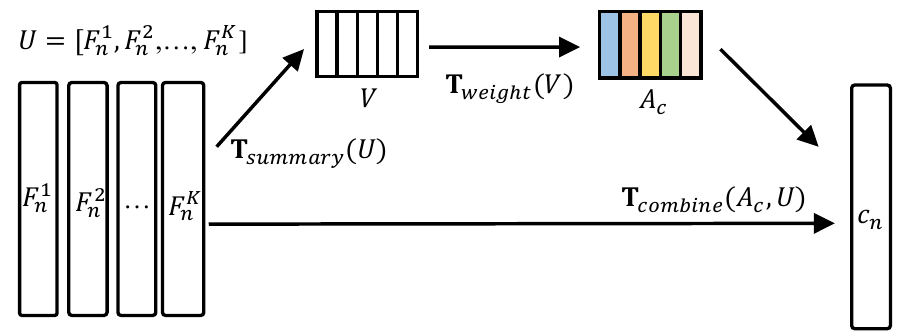}
   \caption{The architecture of the adaptive instance revaluing network. It takes a set of instance-level representations as input, produces the relative significance of the corresponding instance, and then generates a class representation by weightedly combining them.}
   \label{fig:adaptive}
\end{figure}

\subsection{Adaptive Instance Revaluing Network}

After learning the instance-level representation, one crucial problem of the metric-based paradigm for few-shot visual recognition is learning effective class-level representations from few support instance-level representations. Previous works treat all support instances equally when generating class-level representation~\cite{snell2017prototypical,2020A}. However, they ignore the instance diversity, and hence the resulting representation may be biased. For example, some support instances may contain too much background or unrelated concepts information. Therefore, when learning the class-level representation, different instances should have different contributions. In light of the above analysis, we design an adaptive instance revaluing network (AIRN), which revalues the relative significance of all support instances in the same class. The class-level representation is obtained by adaptively and weightedly integrating instance-level representations according to their contributions. In this way, we can dynamically increase the importance value of more informative instances and decrease the weight of less informative instances.
%through the AIRN.

The proposed architecture of AIRN is illustrated in Fig.~\ref{fig:adaptive}.
Specifically, given a support class $n$ with $K$-shot instances, when $K>1$, the union set of instance-level representations is denoted as
\begin{equation}
U=[F_{n}^{1},...,F_{n}^{K}],
\end{equation}
where each instance-level representation $F_{n}^{k}= [r_1, ...,r_d]$, $d$ is the feature dimension and $k \in \left \{ 1, ..., K \right \}$.
Inspired by SENet~\cite{hu2018squeeze}, our we first calculate the ``summary statistics" of each instance-level representation and outputs a statistical vector $V=[v_1,...,v_K]$, where each value $v_k$ is calculated as follows,
\begin{equation}
v_{i} = \mathbf{T}_{summary}\left ( F_{n}^{k}\right )=\frac{1}{d}\sum_{t=1}^{d}{r_t},
\end{equation}
%where ${r_t}$ represents the value of position $t$ in the feature $F_{n}^{k}$ and $d$ is the feature dimension of $F_{n}^{k}$.
Then we utilize two fully connected layers to learn the weight of each instance based on the statistical vector $V$. The set of instance weights $A_c$ for class representation learning can be calculated as:
\begin{equation}
A_c=\mathbf{T}_{weight}(V)=\delta (w_4\sigma (w_3\bm{V})),
\end{equation}
where $A_c = [a_1, \ldots, a_K]$ is a vector, and each value represents the relative significance of a corresponding instance, $\delta$ denotes the sigmoid function, $\sigma$ denotes the ReLU function, and $w_3$ and $w_4$ are the parameters of two fully connected layers.

Finally, the class-level representation of class $n$ is given by:
\begin{equation}
c_n =\mathbf{T}_{combine}(A_c,U) = \sum_{k=1}^{K}a_k\cdot F_{n}^{k},
\end{equation}
where $a_k$ is the relative significance of $k$-th instance.

\subsection{Classifier}
Following~\cite{2020A}, we utilize the cosine classifier to classify query instances in the final classification stage. Specifically, given a query instance $q_{n}^{m}$, a classifier is followed by a softmax activation to generate a probability distribution, which is defined as
\begin{equation}
    p(y=n|c_n,q_{n}^{m})=\frac{\exp(w_n^{T}\cdot \hat{F_{n}^{m}})} {\sum_{i=1}^{N}\exp(w_i^{T}\cdot \hat{F_{n}^{m}})},
\end{equation}
and the predicted label of the query instance is
\begin{equation}
    \hat{y}=\underset{n \in \left \{ 1,...,N \right \} }{\arg\max}(w_n^{T}\cdot \hat{F_{n}^{m}}),
\end{equation}
where $F_{n}^{m}$ is the instance-level representation of $q_{n}^{m}$,
%$\hat{F_{n}^{m}}= \frac{F_{n}^{m}}{\left \| F_{n}^{m} \right \|_{2}}$ is the normalized feature of $F_{n}^{m}$,
and $w_n=\frac{c_n}{\left \| c_n \right \|_{2}}$ represents the weight parameter of class $n$ in the cosine classifier.

\subsection{Optimization Loss}
After predicting the labels of the instances in the query set $\mathcal{Q}$, we have the following classification loss on the query set:
\begin{equation}
L_{cls} =\sum_{( q_{n}^{m},y_{n}^{m} \in \mathcal{Q})} log(\hat{y}=y_{n}^{m}|q_{n}^{m},\left \{ c_n \right \}),
\end{equation}
where $L_{cls}$ is the cross-entropy loss, $y_{n}^{m}$ and $\hat{y}$ are the ground truth label and predicted label for the query instance $q_{n}^{m}$.

Meanwhile, in order to facilitate the learning of class-level representation, one solution is to decrease the similarities between the class-level representation of different classes and increase the similarities between the support instances and its own class-level representation. For example,~\cite{goldblum2020unraveling} exploit the intra-class to inter-class variance-ratio to measure feature clustering and minimize the feature clustering loss to reduce intra-class variation among features during the training. This helps increase the inter-class distance and decrease the intra-class distance in the representation space, and enables the support instances to be more similar in the same class and dissimilar to other categories. Then the query instances can be easily classified. To achieve this goal, we design two novel structural losses, \textit{i.e.}, intra-class instance clustering loss denoted as $L_{intra}$ and inter-class representation distinguishing loss denoted as $L_{inter}$, to further refine the class-level representation, i.e.,
\begin{equation}
L_{intra} = \sum_{(x_n^{k},y_n^{k} \in S)}log(\hat{y}=y_n^{k}|x_n^{k},\left \{ c_n \right \}),
\end{equation}
\begin{equation}
L_{inter} =\sum_{i\neq j} {\hat{c_{i}}^{T}\hat{c_{j}}},
\end{equation}
where $\hat{c_{i}}=\frac{c_i}{\left \| c_i \right \|_{2}}$ and $\hat{c_{j}}=\frac{c_j}{\left \| c_j \right \|_{2}}$.
%are normalization values of $c_{i}$ and $c_{j}$, respectively.

The overall objective function is a combination of $L_{cls}$, $L_{intra}$ and $L_{inter}$, \textit{i.e.},
\begin{equation}
L_{joint}=L_{cls} + \lambda_1 L_{intra} + \lambda_2 L_{inter},
\end{equation}
where $\lambda_1$ and $\lambda_2$ are trade-off hyper-parameters. The training procedure and the inference procedure of the ICRL-Net are summarized in Algorithm \ref{alg:training} and Algorithm \ref{alg:inference}, respectively.

\begin{algorithm}[!t]
\caption{The training process of ICRL-Net}
\label{alg:training}
\textbf{Input}: Training set $\mathcal{D}_{train}$
\begin{algorithmic}[1] %[1] enables line numbers
\WHILE{not done}
\STATE Sample a $N$-way $K$-shot task ($\mathcal{S}$,$\mathcal{Q}$) from $\mathcal{D}_{train}$
\FOR{$n$ in ${1,...,N}$}
\FOR{$k$ in ${1,...,K}$}
\STATE Generate $F_{n}^{k}$ for support instance $x_{n}^{k}$ using Eq.~(2)
\ENDFOR
\STATE Generate $c_n$ for class $n$ using Eq.~(3)-Eq.~(6)
\ENDFOR
\FOR{$n$ in ${1,...,N}$}
\FOR{$m$ in ${1,...,M}$}
\STATE Generate $F_{n}^{m}$ for instance $q_{n}^{m}$ using Eq.~(2)
\STATE Predict $\hat{y}$ for query instance $q_{n}^{m}$ using Eq.~(8)
\ENDFOR
\ENDFOR
\STATE Compute the loss function $L_{joint}$ using Eq.~(12)
\STATE Update the parameters of ICRL-Net with $\bigtriangledown_{L_{joint}}$ using SGD
\ENDWHILE
\end{algorithmic}
\end{algorithm}

\begin{algorithm}[!t]
\caption{The inference process of ICRL-Net}
\label{alg:inference}
\textbf{Input}: Testing set $\mathcal{D}_{test}$\\
\textbf{Require:} The trained ICRL-Net
\begin{algorithmic}[1] %[1] enables line numbers
%\FORALL{epidode = 1, ... , TESTepisode}
\STATE Sample a $N$-way $K$-shot task ($\mathcal{S}$,$\mathcal{Q}$) from $\mathcal{D}_{test}$
\FOR{$n$ in ${1,...,N}$}
\FOR{$k$ in ${1,...,K}$}
\STATE  Generate $F_{n}^{k}$ for support instance $x_{n}^{k}$ using Eq.~(2)
\ENDFOR
\STATE Generate $c_n$ for each class $n$ using Eq.~(3)-Eq.~(6)
\ENDFOR
\FOR{$n$ in ${1,...,N}$}
\FOR{$m$ in ${1,...,M}$}
\STATE Generate $F_{n}^{m}$ for query instance $q_{n}^{m}$ using Eq.~(2)
\STATE Predict $\hat{y}$ for query instance $q_{n}^{m}$ using Eq.~(8)
\ENDFOR
\ENDFOR
\STATE Compute the predict accuracy for each episode (task)
%\ENDFOR
\end{algorithmic}
\end{algorithm}

\begin{table*}[!t]
\small
 \centering
 \caption{Comparisons with other state-of-the-art methods on miniImageNet and tieredImageNet. We use the officially provided results of all the other methods. The mean accuracy($\%$) of the proposed method ICRL-Net is obtained by over 600 testing episodes followed by the 95$\%$ confidence intervals($\%$). For each setting, the best result is highlighted. `-' : not reported.}
 \begin{tabular}{lccccc}
   \toprule[2pt]

   {} & {} & \multicolumn{2}{c}{miniImageNet 5-way} & \multicolumn{2}{c}{tieredImageNet 5-way} \\

   \specialrule{0em}{2pt}{2pt}
   \cline{3-6}
   Model & Backbone & 1-shot & 5-shot & 1-shot & 5-shot\\
    \midrule
     Matching Networks\cite{vinyals2016matching} & ConvNet-4 & 43.56 $\pm$ 0.84 & 55.31 $\pm$ 0.73 & - & - \\
     Prototypical Networks~\cite{snell2017prototypical} & ConvNet-4 &49.42 $\pm$ 0.78 & 68.20 $\pm$ 0.66 & 53.31 $\pm$ 0.89 & 72.69 $\pm$ 0.74 \\
     MAML~\cite{finn2017model} & ConvNet-4 & 48.70 $\pm$ 1.84 & 63.10 $\pm$ 0.92  & 51.67 $\pm$ 1.81 & 70.30 $\pm$ 1.75 \\
     Relation Networks~\cite{sung2018learning} & ConvNet-4  & 50.44 $\pm$ 0.82 & 65.32 $\pm$ 0.70 & 54.48 $\pm$ 0.93 & 71.32 $\pm$ 0.78 \\
     Das et.al~\cite{8935497} & ConvNet-4  & 52.68 $\pm$ 0.51 & 70.91 $\pm$ 0.85 & - & - \\
     wDAE-GNN~\cite{gidaris2019generating} & WRN-28-10 & 62.96 $\pm$ 0.15 & 78.85 $\pm$ 0.10 & 68.18 $\pm$ 0.16 & 83.09 $\pm$ 0.12\\
     LEO~\cite{rusu2018meta} & WRN-28-10 & 61.76 $\pm$ 0.08 & 77.59 $\pm$ 0.12 & 66.33 $\pm$ 0.05 & 81.44 $\pm$ 0.09 \\
     AWGIM~\cite{guo2020attentive} & WRN-28-10 & 63.12 $\pm$ 0.08 & 78.40 $\pm$ 0.11 & 67.69 $\pm$ 0.11 & 82.82 $\pm$ 0.13 \\
     PSST~\cite{chen2021pareto} & WRN-28-10  & 64.16 $\pm$ 0.44 & 80.64 $\pm$ 0.32 & - & - \\
     MTUNet~\cite{wang2021mtunet} & WRN-28-10 & 56.12 $\pm$ 0.43  & 71.93 $\pm$ 0.40  & 62.42 $\pm$ 0.51 & 80.05 $\pm$ 0.46 \\
     MPM~\cite{9159937} & WRN-28-10 & 61.77 & 78.03 & 67.58 & 83.93 \\
     AFHN~\cite{li2020adversarial} & ResNet-18 & 62.38 $\pm$ 0.72 & 78.16 $\pm$ 0.56 & - & - \\
     VI-Net~\cite{luo2021few} & ResNet-18 & 61.05 & 78.60 & - & - \\
    % Dynamic FSL~\cite{gidaris2018dynamic} & ResNet-12 & 56.20 $\pm$ 0.86 & 73.00 $\pm$ 0.64 & - & -\\
     TADAM~\cite{oreshkin2018tadam} & ResNet-12 & 58.50 $\pm$ 0.30 & 76.70 $\pm$ 0.30 & - & - \\
     MTL~\cite{sun2019meta} & ResNet-12 & 61.20 $\pm$ 1.80 & 75.50 $\pm$ 0.80 & - & - \\
     MetaOptNet~\cite{lee2019meta} & ResNet-12 & 62.64 $\pm$ 0.61 & 78.63 $\pm$ 0.46 & 65.99 $\pm$ 0.72 & 81.56 $\pm$ 0.53 \\
     CAN~\cite{hou2019cross} & ResNet-12 & 63.85 $\pm$ 0.48 & 79.44 $\pm$ 0.34 & 69.89 $\pm$ 0.51 &84.23 $\pm$ 0.37 \\
     METANAS~\cite{elsken2020meta} & ResNet-12 & 61.70 $\pm$ 0.30 & 78.80 $\pm$ 0.20 & - & - \\
     DSN~\cite{2020Adaptive} & ResNet-12 & 62.64 $\pm$ 0.66 &78.83 $\pm$ 0.45  & 66.22 $\pm$ 0.75 &82.79 $\pm$ 0.48 \\
     RFS~\cite{tian2020rethinking} & ResNet-12  & 62.02 $\pm$ 0.63& 79.64 $\pm$ 0.44&  69.74 $\pm$ 0.72 & 84.41 $\pm$ 0.55 \\
     SLA-AG~\cite{lee2020self} & ResNet-12 & 62.93 $\pm$ 0.63 & 79.63 $\pm$ 0.47 & - &  - \\
     FEAT~\cite{ye2020few} & ResNet-12 &	\textbf{66.78} &	82.05&	70.80	&84.79 \\
     DeepEMD~\cite{zhang2020deepemd}&	ResNet12&	65.9 $\pm$ 0.82&	\textbf{82.41 $\pm$ 0.56}	& \textbf{71.16 $\pm$ 0.87}	& \textbf{86.03 $\pm$ 0.58} \\
     MAN~\cite{li2021multi} & ResNet-12 & 61.70 $\pm$ 0.47 & 78.42 $\pm$ 0.34 & 65.99 $\pm$ 0.51 & 81.97 $\pm$ 0.37 \\
     ConstellationNet~\cite{xu2021attentional} & ResNet-12 & 64.89 $\pm$ 0.23 & 79.95 $\pm$ 0.17 & - & - \\
%     Koo et al.~\cite{koo2021improving} &  ResNet-12 & 64.86 $\pm$ 0.21 & 81.25 $\pm$ 0.14 & 68.87 $\pm$ 0.23 &  85.07 $\pm$ 0.16\\
   %  DCAP & ResNet-12 & 65.20 $\pm$ 0.67 &  80.93 $\pm$ 0.53 & 70.15 $\pm$ 0.74 & 85.33 $\pm$ 0.55 \\
     Meta-UAFS~\cite{zhang2020uncertainty} & ResNet-12 & 64.22 $\pm$ 0.67 & 79.99 $\pm$ 0.49 & 69.13 $\pm$ 0.84 & 84.33 $\pm$ 0.59 \\
     Meta-Baseline~\cite{2020A} & ResNet-12 & 63.17 $\pm$ 0.23 & 79.26 $\pm$ 0.17 & 68.62 $\pm$  0.27 & 83.29 $\pm$ 0.18 \\
   \midrule
     Baseline & ResNet-12 & 62.71 $\pm$ 0.77& 79.34 $\pm$ 0.56&	68.30 $\pm$ 0.80	&83.57 $\pm$ 0.63 \\
     ICRL-Net (OURS) & ResNet-12 & 65.55 $\pm$ 0.79 & 81.87 $\pm$ 0.51 & 70.56 $\pm$ 0.91 & 85.62 $\pm$ 0.64\\
     \bottomrule[2pt]
 \end{tabular}
\label{tab:mini}
\end{table*}

\begin{table*}[!t]
\small
 \centering
  \caption{A comparison of different methods on CIFAR-FS and FC100. We use the officially provided results of all the other methods. The mean accuracy($\%$) over 600 testing episodes is reported followed by the 95$\%$ confidence intervals($\%$). For each setting, the best result is highlighted. `-' : not reported.}
 \begin{tabular}{lccccc}
   \toprule[2pt]
   {} & {} & \multicolumn{2}{c}{CIFAR-FS 5-way} & \multicolumn{2}{c}{FC100 5-way} \\

   \specialrule{0em}{2pt}{2pt}
   \cline{3-6}
   Model & Backbone & 1-shot & 5-shot & 1-shot & 5-shot\\
    \midrule
     MAML~\cite{vinyals2016matching} & ConvNet-4 & 58.9 $\pm$ 1.9 & 71.5 $\pm$ 1.0 & - & - \\
     Prototypical Networks~\cite{snell2017prototypical} & ConvNet-4 & 55.5 $\pm$ 0.7 & 72.0 $\pm$ 0.6 & 35.3 $\pm$ 0.6 & 48.6 $\pm$ 0.6 \\
     Relation Networks~\cite{sung2018learning} & ConvNet-4 & 55.0 $\pm$ 1.0 & 69.3 $\pm$ 0.8 & - & - \\
     TADAM~\cite{oreshkin2018tadam} & ResNet-12  & - & - & 40.1 $\pm$ 0.4 & 56.1 $\pm$ 0.4 \\
     MetaOptNet~\cite{lee2019meta} & ResNet-12 & 72.0 $\pm$ 0.7 & 84.3 $\pm$ 0.5 & 41.1 $\pm$ 0.6 & 55.5 $\pm$ 0.6 \\
     DSN~\cite{2020Adaptive} & ResNet-12 & 72.3 $\pm$ 0.8 & 85.1 $\pm$ 0.6 & - & - \\
     SLA-AG~\cite{lee2020self} & ResNet-12 & 73.5 $\pm$ 0.7 & 86.7 $\pm$ 0.5 & 42.2 $\pm$ 0.6 & 59.2 $\pm$ 0.5 \\
     MABAS~\cite{kim2020model} & ResNet-12 & 73.5 $\pm$ 0.9	& 85.7 $\pm$ 0.6 & 42.3 $\pm$ 0.7 & 58.2 $\pm$ 0.7 \\
     ConstellationNet~\cite{xu2021attentional} & ResNet-12 & 75.4 $\pm$ 0.2 & 86.8 $\pm$ 0.2 & 43.8 $\pm$ 0.2 & 59.7 $\pm$ 0.2 \\
     Meta-UAFS~\cite{zhang2020uncertainty} & ResNet-12 & 74.1 $\pm$ 0.7 & 85.9 $\pm$ 0.4 & 42.0 $\pm$ 0.6 & 57.4 $\pm$ 0.4 \\
     \midrule
     Baseline & ResNet-12 & 73.6 $\pm$ 0.6 & 85.4 $\pm$ 0.5 & 41.2  $\pm$ 0.6 & 57.1 $\pm$ 0.4 \\
     ICRL-Net (OURS) & ResNet-12  & \textbf{75.8 $\pm$ 0.7} & \textbf{87.3 $\pm$ 0.5} & \textbf{44.5 $\pm$ 0.6} & \textbf{59.8 $\pm$ 0.6}\\
     \bottomrule[2pt]
 \end{tabular}
\label{tab:cifar}
\end{table*}

\section{Experiments} \label{sec:Experiments}
To evaluate the effectiveness of the ICRL-Net, we conducted extensive experiments on four publicly available and widely used few-shot visual recognition benchmarks,~\textit{i.e.}, miniImageNet, tieredImageNet, CIFAR-FS, and FC100 datasets. In this section, we first introduce dataset details and experimental settings, and then comparisons with the state-of-the-art approaches. Finally, comprehensive ablation study is conducted to verify effectiveness of different components.

\subsection{Dataset}
\subsubsection{MiniImageNet}

The miniImageNet dataset was initially proposed by~\cite{vinyals2016matching}, is a standard benchmark for few-shot visual recognition. MiniImageNet is a subset randomly sampled from the ImageNet~\cite{russakovsky2015imagenet} dataset. MiniImageNet includes a total number of 100 classes and 600 images per class. We follow the split strategy proposed in~\cite{ravi2016optimization} to split all 100 classes into three subsets. One subset that contains 64 classes is used for training. The other two subsets used for validation and testing include 16 and 20 classes, respectively. All images are resized to 84$\times$84.

\subsubsection{TieredImageNet}
The tieredImageNet dataset was proposed by~\cite{ren2018meta}. TieredImageNet is a relatively large subset sampled from ImageNet~\cite{russakovsky2015imagenet} dataset and consists of 608 classes that can be grouped into 34 high-level categories. The tieredImageNet dataset is split into three subsets: a training set, a validation set, and a testing set with 20, 6, and 8 high-level categories. The corresponding numbers of classes are 351, 97, and 160, respectively. All images are of size 84$\times$84.

\subsubsection{CIFAR-FS}
The CIFAR-FS dataset~\cite{bertinetto2018meta} is a recently proposed few-shot visual recognition benchmark, consisting of all 100 classes from CIFAR-100~\cite{krizhevsky2010cifar}. All images on these datasets are 32 $\times$ 32, and the number of images per class is 600. Following previous works~\cite{lee2019meta,2020Adaptive}, we divide the whole dataset into 64, 16, and 20 classes for training, validation, and testing, respectively.

\subsubsection{FC100}
The FC100 dataset~\cite{oreshkin2018tadam} is another benchmark derived from CIFAR-100~\cite{krizhevsky2010cifar}.
There are 60 classes from 12 different superclasses for training, 20 classes from 4 different superclasses for validation, and 20 classes from 4 different superclasses for testing. Similar to the CIFAR-FS dataset, every class has 600 images of size 32 $\times$ 32.

\subsection{Implementation Details}

\subsubsection{Pre-training Process} Following the previous works~\cite{2020A,ye2020few}, we apply an additional pre-training phase to train the backbone network. Thus, the backbone network, appended with a linear layer, is trained to classify all training classes (e.g., 64 classes in the miniImageNet) based on the cross-entropy loss. We follow the conventional deep learning pipeline and divide each training class into two parts: model training and validation. In this stage, the learning rate is set to 0.1. For all datasets, random resized crop and horizontal flip data augmentations are used for model optimization. The classification performance over features of sampled 1-shot tasks from the model validation split is evaluated to select the best pre-trained model, whose weights are then used to initialize the backbone network in the meta-training phase.

\subsubsection{Meta-training Process} We follow the episodic training strategy~\cite{vinyals2016matching} to train the ICRL-Net in the meta-training phase. Specifically, the pre-trained backbone network is finetuned at a learning rate of 0.001. Meanwhile, other parts of ICRL-Net are initialized randomly and optimized with a learning rate of 0.01. We set the total number of the training epoch to be 200, and each epoch contains 100 episodes. All experiments are implemented in PyTorch on an Ubuntu server with a single NVIDIA V100 GPU card. We adopt SGD with a Nesterov momentum of 0.9 and weight decay of 0.0005 for model optimization in both pre-training and meta-training. Moreover, the learning rate decay is set to 0.1 in two in both pre-training and meta-training. We adopt the same data augmentation (i.e., random horizontal flip and random resized crop) when dealing with all datasets during meta-training. More implementation details can be found in the supplementary material.

\subsection{Performance Comparison}

We compared the proposed method with state-of-the-art methods, such as MAML~\cite{finn2017model}, LEO~\cite{rusu2018meta}, MTL~\cite{sun2019meta}, Matching Networks~\cite{vinyals2016matching}, Prototypical Networks~\cite{snell2017prototypical}, Relation Networks~\cite{sung2018learning}, CAN~\cite{hou2019cross}, Meta-Baseline~\cite{2020A}, and DSN~\cite{2020Adaptive}.

%Note that optimization-based methods are essentially different from metric-based methods in two respects. The first aspect is that an additional parameterized meta-learner is usually learned in optimization-based methods, while the metric-based methods do not. The second aspect is that optimization-based methods will fine-tune the model (or classifier) in new tasks to obtain the final classification results during the testing process, while metric-based methods do not require fine-tuning.

\subsubsection{Performance on miniImageNet and tieredImageNet}

Table~\ref{tab:mini} shows the performance of all comparison methods on miniImageNet and tieredImageNet. The best results are highlighted in boldface.

%The performance of CAN and DSN in Table~\ref{tab:mini} are not the best as reported in their original papers. This is because CAN exploits a transductive strategy that all query images are used simultaneously during inference, and DSN considers semi-supervised few-shot settings to refine each class's prototype by leveraging massive unlabeled data. For a fair comparison, we use the inductive setting without using unlabeled data for all compared methods, including CAN and DSN.
On the miniImagnet dataset, we observe that the proposed method ICRL-Net can achieve comparable performance compared with state-of-the-art methods under both 5-way 1-shot and 5-shot settings and achieves the accuracies with 65.55\% and 81.87\% on 1-shot and 5-shot, respectively. Compared with several state-of-the-art approaches, such as PSST~\cite{chen2021pareto}, ConstellationNet~\cite{xu2021attentional}, and Meta-Baseline~\cite{2020A}. We observe that the proposed ICRL-Net achieves the best performance compared to these state-of-the-arts algorithms. Note that our baseline is the same as Meta-Baseline~\cite{2020A}, and they have a similar performance. Compared to Meta-Baseline, ICRL-Net brings significant improvements and improves 1-shot accuracy by 2.38\% and 5-shot accuracy by 2.61\%. Moreover, our approach has such an improvement attributed to the adaptive instance revaluing strategy that facilitates learning optimal class representations. The improvements demonstrate that the necessity of considering the instance diversity for improving the class representation learning.

On the tieredImageNet dataset, it is worth noting that the principle of dividing the training set and testing set is according to the disjoint sets, where the similarity of classes in each disjoint set is relatively high. Thus, it is more difficult to distinguish the categories in the training set and testing set. Nevertheless, on the tieredImageNet dataset, we observe that the proposed ICRL-Net still achieves comparable performance compared to other state-of-the-art algorithms. For example, on tiredImageNet, the 5-way 1-shot and 5-shot accuracies of FEAT are 70.80\% and 84.79\%, respectively, and the accuracies of ICRL-Net are 70.56\% and 85.62\%, respectively. Note that our baseline is the same as Meta-Baseline~\cite{2020A}, and they have a similar performance. Compared to Meta-Baseline~\cite{2020A}, our ICRL-Net achieves a significant improvement of 1.94\% and 2.33\% accuracies for 1-shot classification and 5-shot classification, respectively. The improvements also prove the effectiveness of our approach.

\subsubsection{Performance on CIFAR-FS and FC100} Experimental results on CIFAR-FS and FC100 are shown in Table~\ref{tab:cifar}. The best results are highlighted in boldface. Our method also consistently outperforms the other state-of-the-arts methods under both 1-shot and 5-shot settings on the CIFAR-FS and FC100 datasets. For the CIFAR-FS dataset, our method outperforms the sub-optimal method by 0.4\% on 1-shot and 0.5\% on 5-shot. Notice that our ICRL-Net perform much better than the  methods based on the average class representation: MetaOptNet~\cite{lee2019meta} and Meta-Baseline~\cite{2020A}. For the FC100 dataset, our method also improves the classification performance by 0.7\% and 0.1\% under 1-shot and 5-shot settings, respectively. The results show the necessity and effectiveness of learning an optimal class representation.

\subsection{Ablation Study}

In this subsection, we conduct ablation studies to assess the effectiveness of each component of ICRL-Net, including AIRN, the attentional bilinear feature extractor, and the designed loss function. We refer to the supplementary material for more studies.

\subsubsection{Influence of the AIRN}
%In Table~\ref{tab:component}, we report the comparative results of the above components on both datasets under the standard FSL setting. The baseline is the ICRL-Net without using AIRN and BAP. The class representations are directly averaged from a few support instance-level representations of one class. From Table~\ref{tab:component}, we can conclude that: 1) the baseline obtains the worst performance; both AIRN and BAP contribute to the improvements of ICRL-Net. These improvements indicate that it is necessary to consider instance diversity when studying the class-level representation for each class. In order to obtain the best performance, both AIRN and BAP should be considered; 2) performance of the model that exploits BAP when learning instance-level representation is better than that of the model without BAP. This means that the second-order pooling strategy is conducive to learning an instance-level representation that is discriminative and informative; 3) ICRL-Net adopts the adaptive instance revaluation strategy leads to significant improvements on 5-shot tasks. For 1-shot tasks, since each class only has one instance. Therefore, class-level representation is just the instance-level representation, and AIRN does not take effect in 1-shot tasks. This verifies our intuition that the relative significance of the instance in a set should be different when learning class-level representations.

\begin{table}[!t]
\small
 \centering
 \caption{Ablation study of the AIRN.}
 \begin{tabular}{lcc}
   \toprule[2pt]
   {Method} & {miniImageNet 5-shot} & {tieredImageNet 5-shot}\\

   \midrule
     Baseline & 79.34 $\pm$ 0.56 & 83.57 $\pm$ 0.63  \\
     + AIRN  & 80.63 $\pm$ 0.58 & 84.64 $\pm$ 0.57  \\
     \bottomrule[2pt]
 \end{tabular}
\label{tab:component}
\end{table}

\begin{table}[!t]
 \centering
 \small
 \caption{Ablation study of generating weights to aggregate $K$-shot instances.}
 \begin{tabular}{lcc}
   \toprule[2pt]
   {Method} & {miniImageNet 5-shot} & {tieredImageNet 5-shot}\\

   \midrule
   Similarity-type & 79.25 $\pm$ 0.54 & 84.02 $\pm$ 0.60 \\
   Attention-type &	79.89 $\pm$ 0.56 &	84.27 $\pm$ 0.62 \\
    AIRN & 80.63 $\pm$ 0.58 & 84.64 $\pm$ 0.57  \\
     \bottomrule[2pt]
 \end{tabular}
\label{tab:weight}
\end{table}

\begin{table*}[!t]
\small
 \centering
 \caption{A comparison of different variants of our bilinear feature extractor with the extractor based on naive bilinear pooling~\cite{lin2015bilinear}. In `Model-1', global average pooling (instead of attention pooling) is adopted in the feature extractor of our ICRL-Net; In `Model-2', attention pooling is removed from our extractor; In `Model-3', the two $1 \times 1$ convolutional layers is removed from our extractor; `Model-4' is to utilize only one $1 \times 1$ convolutional layer to transform the features in our extractor; Finally, `Model-5' is to utilize naive bilinear pooling~\cite{lin2015bilinear} in the feature extractor.
 }
 \begin{tabular}{lcccc}
   \toprule[2pt]
   {} & \multicolumn{2}{c}{miniImageNet 5-way} & \multicolumn{2}{c}{tieredImageNet 5-way}\\
   \specialrule{0em}{2pt}{2pt}
   \cline{2-5}
    Method & 1-shot & 5-shot & 1-shot & 5-shot\\
    \midrule
     Model-1 & 63.86 $\pm$ 0.80 & 81.23 $\pm$ 0.50 & 69.50 $\pm$ 0.82 & 84.94 $\pm$ 0.62\\
     Model-2 & 64.86 $\pm$ 0.81 & 81.67 $\pm$ 0.54 & 70.13 $\pm$ 0.83 & 85.33 $\pm$ 0.66 \\
     Model-3 & 64.62 $\pm$ 0.84 & 81.56 $\pm$ 0.53 & 70.22 $\pm$ 0.84 & 85.10 $\pm$ 0.60 \\
     Model-4 & 64.30 $\pm$ 0.77	& 81.25 $\pm$ 0.52 & 69.86 $\pm$ 0.80 &	85.14 $\pm$ 0.63\\
     Model-5 & 64.05 $\pm$ 0.76	& 80.78 $\pm$ 0.56 & 69.44 $\pm$ 0.86 &	84.76 $\pm$ 0.58 \\
    ICRL-Net & 65.55 $\pm$ 0.79 & 81.87 $\pm$ 0.51 & 70.56 $\pm$ 0.91 & 85.62 $\pm$ 0.64\\
     \bottomrule[2pt]
 \end{tabular}
\label{tab:pooling}
\end{table*}

\begin{table*}[htbp]
\small
 \centering
 \caption{The influence of the joint loss function. We compared the joint loss with other loss functions, including the original classification loss function $L_{cls}$ (i.e., cross entropy loss), the $L_{cls}$ + $L_{inter}$ loss function, and the $L_{cls}$ + $L_{intra}$ loss function. }
 \begin{tabular}{lcccc}
   \toprule[2pt]

   {} & \multicolumn{2}{c}{miniImageNet 5-way} & \multicolumn{2}{c}{tieredImageNet 5-way}\\

   \specialrule{0em}{2pt}{2pt}
   \cline{2-5}
    Loss Function & 1-shot & 5-shot & 1-shot & 5-shot \\
    \midrule
     $L_{cls}$ & 64.32 $\pm$ 0.80 & 81.12 $\pm$ 0.52 & 69.62 $\pm$ 0.84 & 85.07 $\pm$ 0.61 \\
     $L_{cls}$ + $L_{inter}$  & 65.26 $\pm$ 0.78 & 81.43 $\pm$ 0.56 & 70.32 $\pm$ 0.91 & 85.38 $\pm$ 0.61 \\
     $L_{cls}$ + $L_{intra}$  & 65.03 $\pm$ 0.80 & 81.50 $\pm$ 0.58 & 69.87 $\pm$ 0.84 & 85.24 $\pm$ 0.62  \\
     $L_{joint}$ & 65.55 $\pm$ 0.79  & 81.87 $\pm$ 0.51 & 70.56 $\pm$ 0.91 & 85.62 $\pm$ 0.64 \\
     \bottomrule[2pt]
 \end{tabular}
\label{tab:loss}
\end{table*}

In our baseline, the class-level representation is generated by directly averaging a few support instance representations, and the performance is similar to Meta-Baseline~\cite{2020A}. Our ICRL-Net improves this baseline by designing the AIRN module, the attentional bilinear feature extractor, and novel loss function. To evaluate the effectiveness of the proposed AIRN, we directly add this module on the baseline. Table~\ref{tab:component} reports the comparison results under the 1-shot and 5-shot settings. From the results, we observe that utilizing AIRN leads to significant improvements on 5-shot tasks compared with the baseline. These improvements verify our intuition that the relative importance of few-shot support instances in the same class should be different when learning class-level representations. For 1-shot tasks, since there is only one support instance per class. Therefore, class-level representation is just the instance-level representation, and AIRN does not take effect in 1-shot tasks.

To further demonstrate effectiveness of the proposed AIRN, we add some experiments that adopt other approaches to aggregate $K$-shot instances, such as similarity-type weights, attention-type weights.
%For example, we adopt similarity-types weights to aggregate $K$-shot support instances. In specific, given $K$-shot support instances of each class $(x_1,¡­,x_K)$, we first calculate the class prototype $\mu =\frac{1}{K}  {\textstyle \sum_{i=1}^{K}}\hat{x_i}$  following the Prototypical Network, and then compute the similarity $a_i = \left \langle \hat{x_i}, \mu  \right \rangle =\frac{\hat{x_i}\cdot \mu  } {\left \| \hat{x_i} \right \| \cdot  \left \| \mu  \right \|  } $) between each support instance and the class prototype as the weight to aggregate the $K$-shot instances. The similarity-based class prototype is updated by the formulation $c = {\textstyle \sum_{i=1}^{K}} a_i \cdot \hat{x_i}$.
The experimental results are shown in Table~\ref{tab:weight}. The AIRN module allows the model to pay more attention to the most informative support instance features while suppressing those unimportant support instance features, and has the best performance when learning class-level representations.

\subsubsection{Influence of the Attention Bilinear Feature Extractor}
To verify the effectiveness of the attentional bilinear feature extractor, we develop various variants of ICRL-Net to perform the process from the instance features to instance-level representations. In `Model-1', global average pooling (instead of attention pooling) is adopted in the feature extractor of our ICRL-Net; In `Model-2', attention pooling is removed from our extractor; In `Model-3', the two $1 \times 1$ convolutional layers is re-moved from our extractor; `Model-4' is to utilize only one $1 \times 1$ convolutional layer to transform the features in our extractor; Finally, `Model-5' is to utilize naive bilinear pooling~\cite{lin2015bilinear} in the feature extractor. Table~\ref{tab:pooling} shows the comparison results under 1-shot and 5-shot settings. The results indicate that the attentional bilinear feature extractor yields better results than other variants. Therefore, it can conclude that the attention bilinear strategy is conducive to learning a discriminative and informative instance-level representation.

\subsubsection{Influence of the Joint Loss Functions}
To demonstrate the effectiveness of the designed loss function, we compare the joint loss with three loss functions, including the original classification loss function $L_{cls}$, the $L_{cls}$ + $L_{inter}$ loss function, and the $L_{cls}$ + $L_{intra}$ loss function. The results are shown in Table~\ref{tab:loss}. Compared with other loss functions, the designed loss function achieves the highest accuracy in 5-way 1-shot and 5-way 5-shot tasks. Moreover, the accuracies of using the $L_{cls}$ + $L_{inter}$ and $L_{cls}$ + $L_{intra}$ are higher than the original classification loss function $L_{cls}$, indicating that the proposed loss function is efficient for the representation learning of FSL.

\subsubsection{Comparison on Different K-shot Settings}
\begin{figure}[!t]
\centering
\subfigure[]{
    \begin{minipage}[t]{0.9\linewidth}
        \centering
        \includegraphics[width=0.9\linewidth]{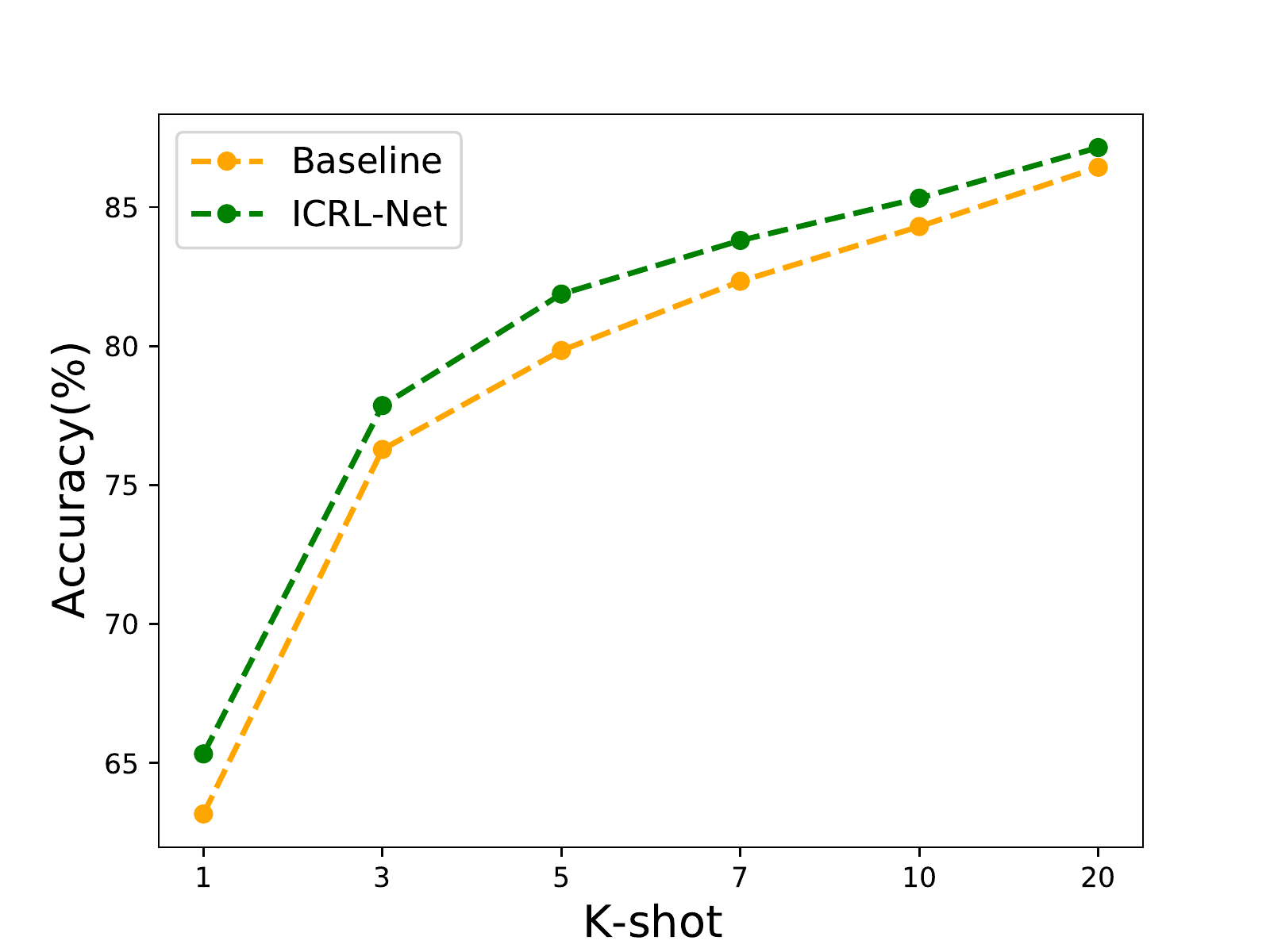}
    \end{minipage}
}
\subfigure[]
{
 	\begin{minipage}[t]{0.9\linewidth}
        \centering
        \includegraphics[width=0.9\linewidth]{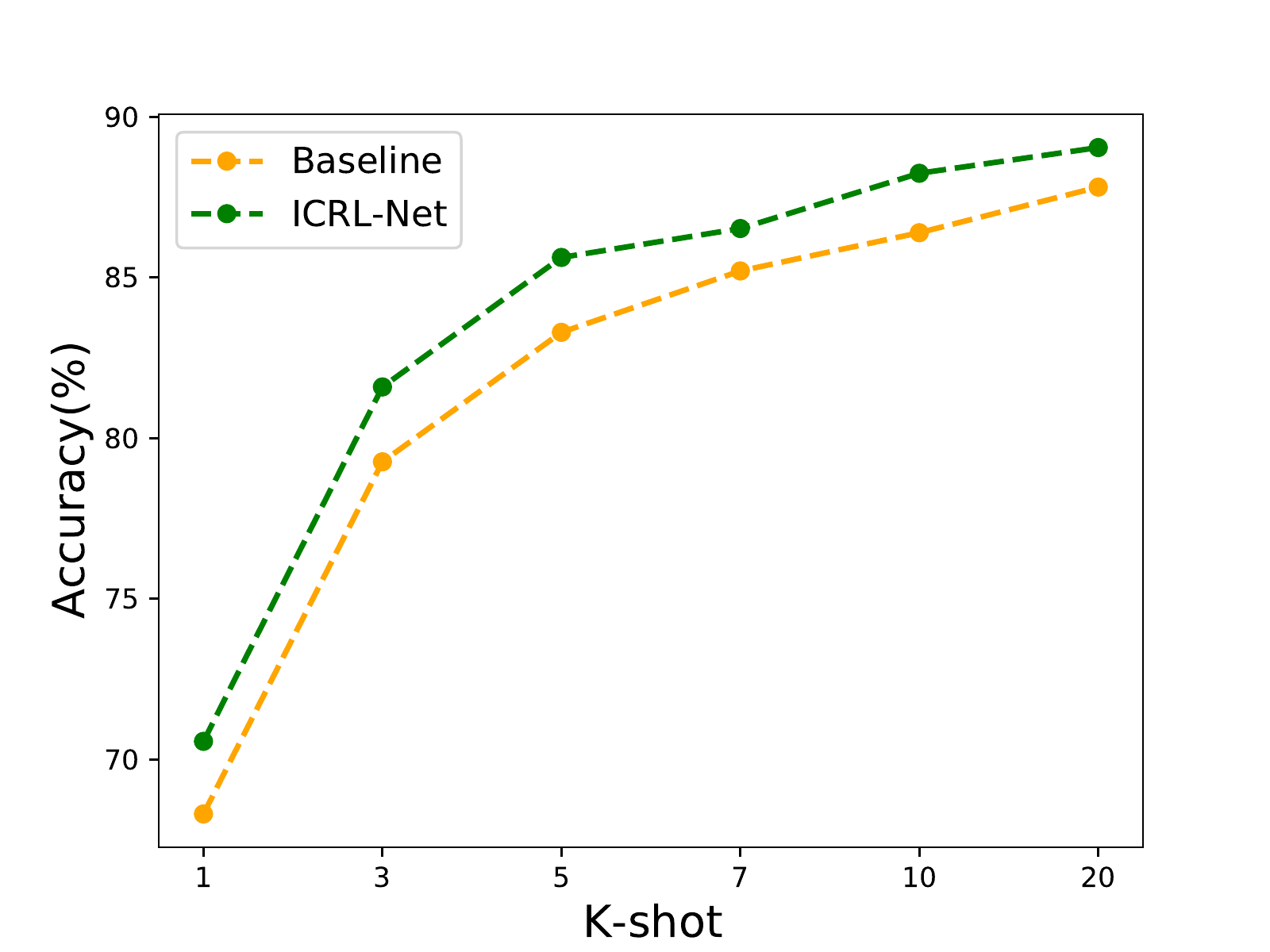}
    \end{minipage}
}
\caption{The results w.r.t. different K-shot settings on miniImageNet and tieredImageNet. (a): 5-way classification on miniImageNet. (b): 5-way classification on tieredImageNet.}
\label{fig:shotnumber}
\end{figure}

To further demonstrate the effectiveness of the adaptive instance revaluation strategy, we compare ICRL-Net with the baseline method under more $K$-shot ($K$=1,3,5,7,10,20) settings. The comparison results are shown in Fig.~\ref{fig:shotnumber}. From the results, we observe that ICRL-Net consistently outperforms the baseline under various $K$-shot settings on miniImageNet and tieredImageNet datasets. The improvements again demonstrate the ability of ICRL-Net and validate the necessity and effectiveness of revaluing the importance of support instances when obtaining the class-level representation.

\begin{figure}[!t]
\centering
\subfigure[]{
    \begin{minipage}[t]{0.9\linewidth}
        \centering
       \includegraphics[width=0.9\linewidth]{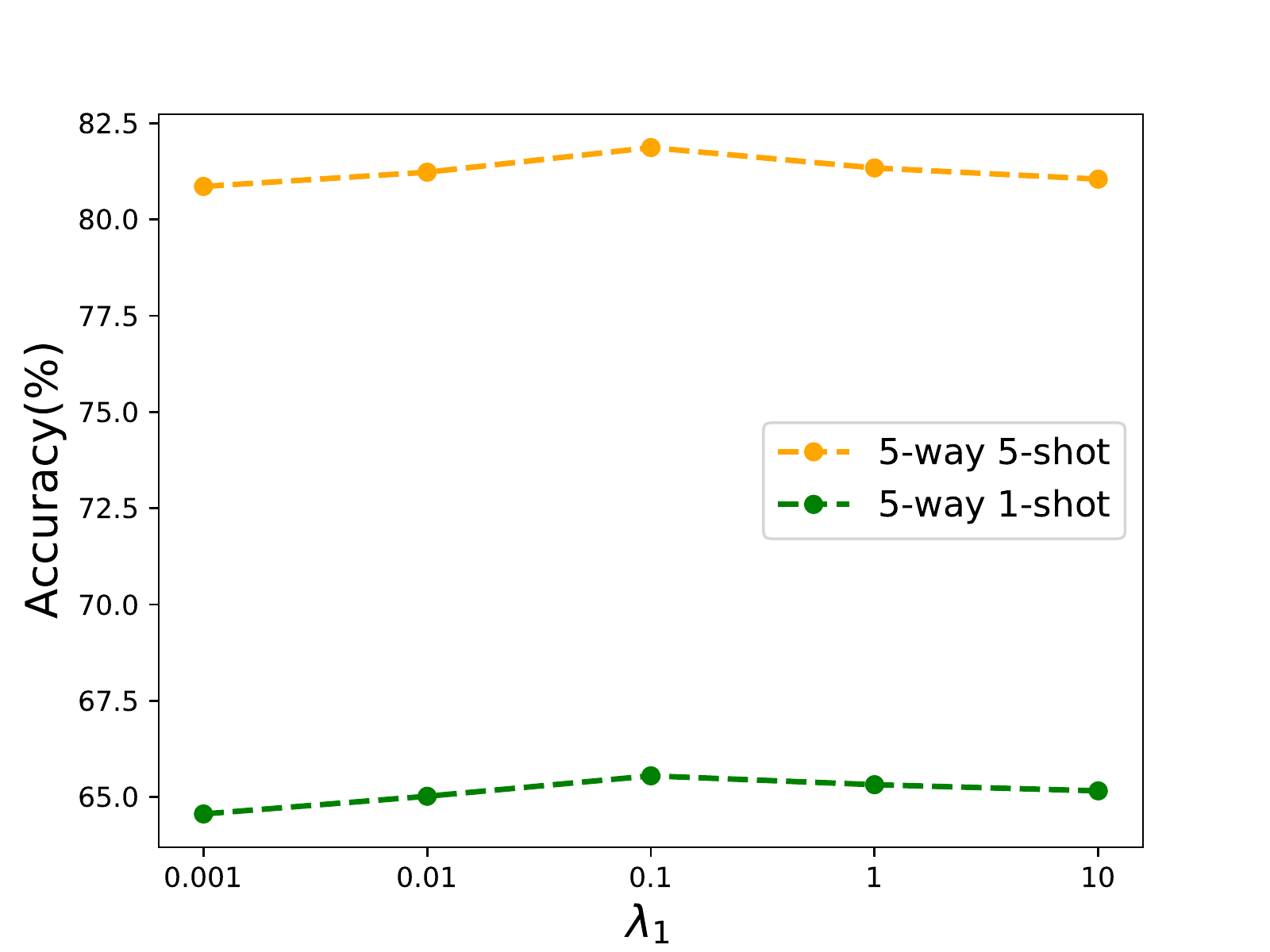}
    \end{minipage}
}
\subfigure[]
{
 	\begin{minipage}[t]{0.9\linewidth}
        \centering
        \includegraphics[width=0.9\linewidth]{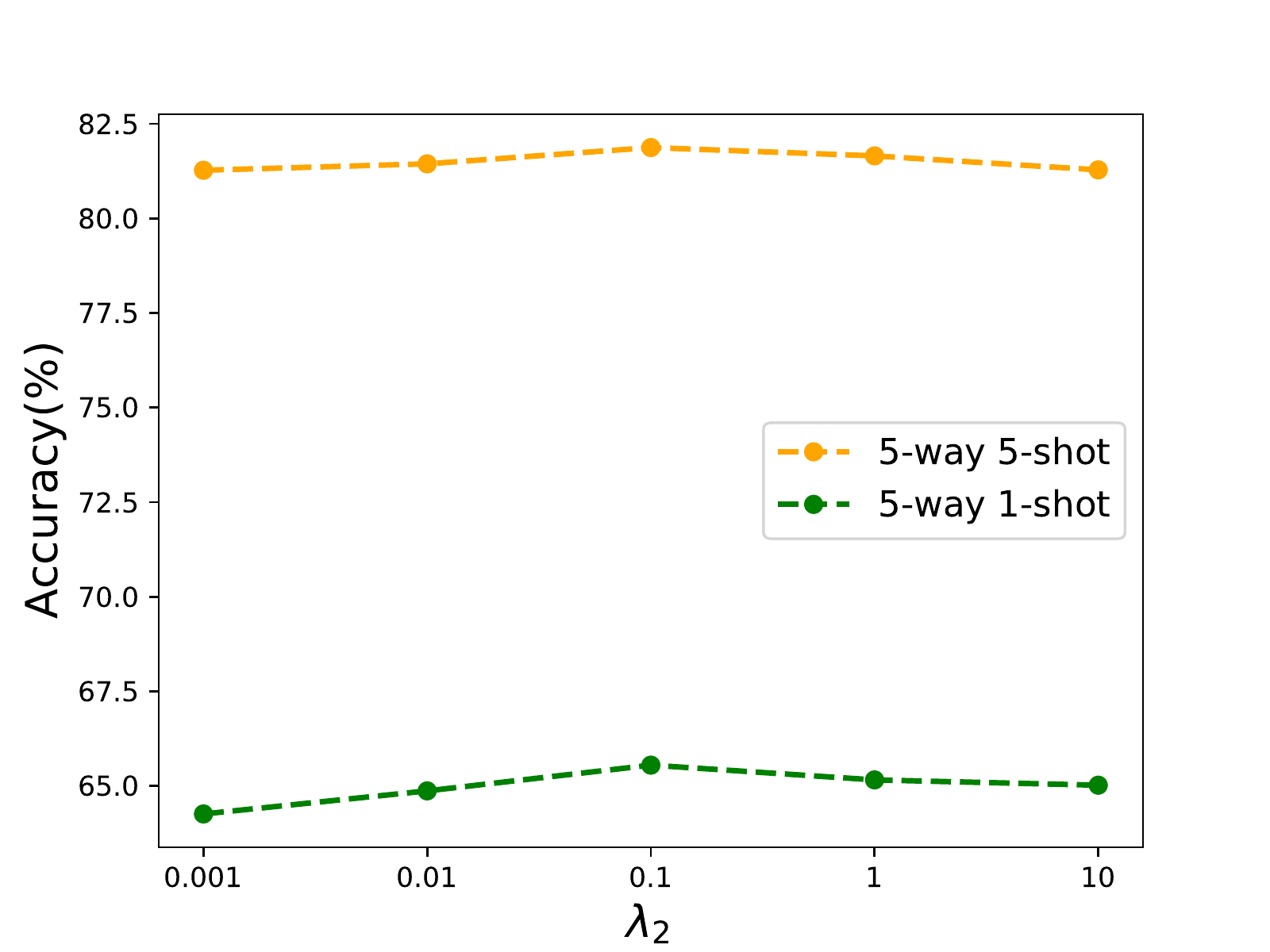}
    \end{minipage}
}
\caption{The results of parameters analysis on miniImageNet.}
\label{fig:balance}
\end{figure}

\subsubsection{Hyper-parameter Analysis}
We also conduct sensitivity analysis for the two trade-off hyper-parameters in our joint loss. The experiments are performed by fixing the value of one hyper-parameter and then changing the value of another hyper-parameter. In specific, when analyzing hyper-parameter $\lambda_1$, the value of hyper-parameter $\lambda_2$ is set to 0.1. Similarly, when analyzing hyper-parameter $\lambda_2$, the value of hyper-parameter $\lambda_1$ is set to 0.1. The experimental results are shown in Fig.~\ref{fig:balance}. We observe that the best performances are achieved when $\lambda_1 = \lambda_2 =0.1$, and this setting is adopted in all experiments.

%\subsection{Visualization}

\subsection{Visualization of Relative Significance}
\begin{figure*}[t]
    \center
    \includegraphics[width=0.93\linewidth]{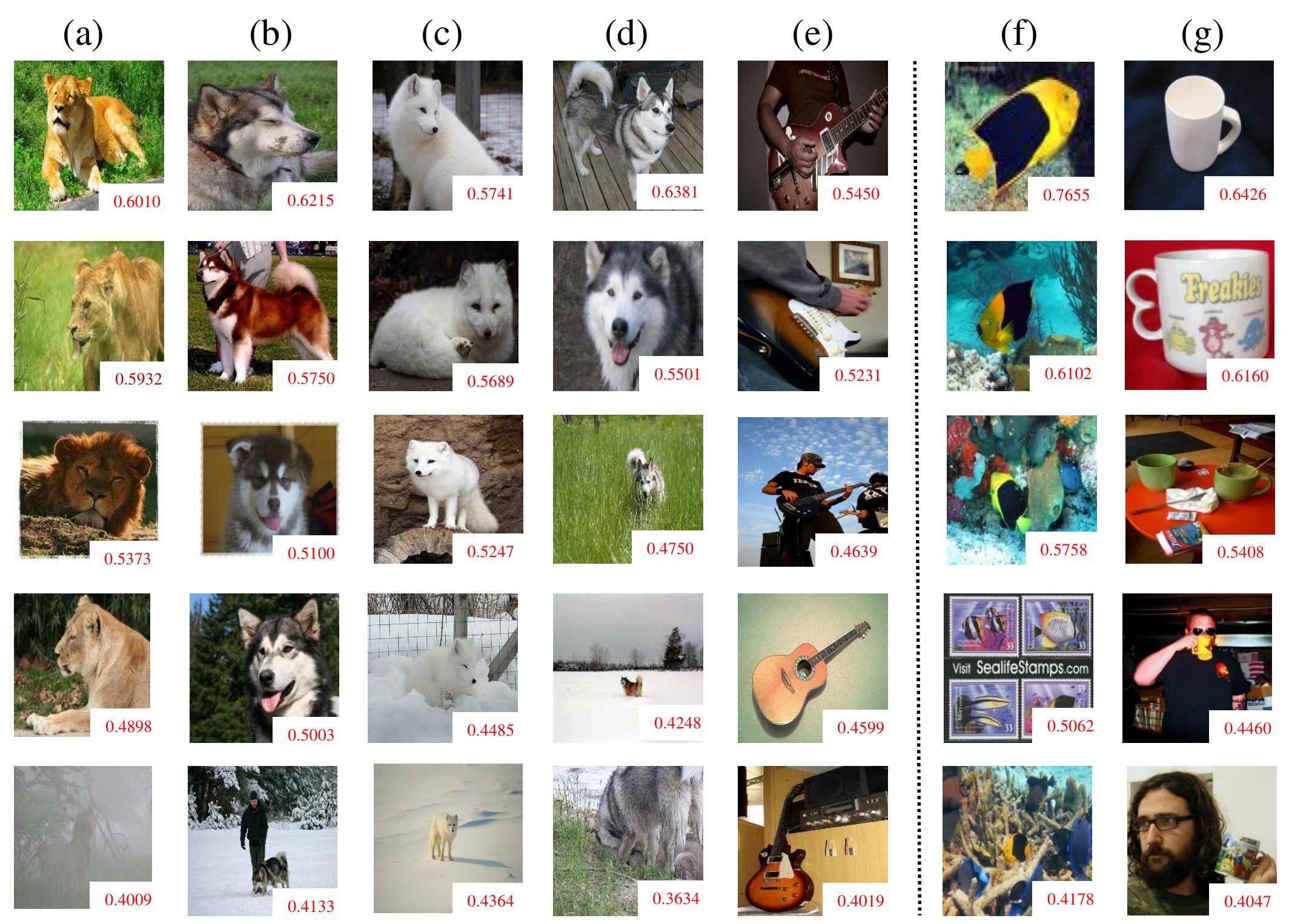}
   \caption{Visualization of the learned importance: (a), (b), (c), (d) and (e) show support instances sampled from miniImageNet, and each class contains five support instances; (f) and (g) show support instances sampled from tieredImageNet. Each column of support instances is sorted by relative importance.}
   \label{fig:weight}
\end{figure*}

We visualize the values of the relative significance of support instances to verify the effectiveness of the proposed method. We randomly selected several group examples from the miniImageNet and tieredImageNet, and each group contains five support instances from the same class. The visualization results of the relative importance are shown in Fig.~\ref{fig:weight}. Each column (group) of support instances is sorted by relative importance. It shows that more informative instances are generally assigned with larger importance weights according to ICRL-Net, except the column (e). This might due to multiple instances contain information related to other concepts, and our method does not accurately express the actual concepts of the instances. Therefore, in the future, we will focus on correctly expressing the target information related to the category when the instance contains multiple targets. We also visualize the learned attention maps in the supplementary material.

\begin{table}[!t]
\small
 \centering
 \caption{5-shot classification accuracy under dataset shift.}
 % \begin{tabular*}{\hsize}{@{}@{\extracolsep{\fill}}lcc@{}}
 \begin{tabular}{lcc}
   \toprule[2pt]
   Model (\%)  & miniImageNet $\rightarrow$  CUB\\
    \midrule
     Linear Classifier~\cite{chen2019closer} & 65.57 $\pm$ 0.70 \\
     Cosine Classifier~\cite{chen2019closer} & 62.04 $\pm$ 0.76 \\
     MetaOptNet-SVM~\cite{lee2019meta} & 54.67 $\pm$ 0.56 \\
     \midrule
     Baseline & 64.86 $\pm$ 0.73 \\
     \textbf{ICRL-Net} & \textbf{67.08 $\pm$ 0.55} \\
     \bottomrule[2pt]
 \end{tabular}
\label{tab:cross-domain}
\end{table}

\subsection{Cross-domain FSL}

Beyond the standard single domain few-shot classification setting, we introduce a more challenging and realistic cross-domain setting. In this new setting, we aim to test the generalization of a trained model across previously unseen domains (datasets) with different data distributions. Cross-domain few-shot learning can better evaluate the model's generalization ability to novel tasks. Following the instructions of ~\cite{chen2019closer}, we conducted a cross-domain few-shot experiment by training the models on the miniImageNet training set and evaluating the model on the CUB200 testing set. This set of experiments is designed to evaluate the performance of different algorithms when the distribution divergence between training and testing sets is large. To fairly compare with other approaches, we adopt the CUB testing split presented in their original work. The comparison results are reported in Table~\ref{tab:cross-domain}, where we can see that the proposed ICRL-Net can outperform other approaches. The cross-domain few-shot settings further demonstrate the effectiveness of the proposed method.

\begin{table}[!t]
  \centering
  \small
  \caption{Transductive FSL on miniImagenet and tieredImageNet. }
  \begin{tabular}{llcc}
   \hline
    Dataset & Method & 1-shot & 5-shot \\
     \hline

      \multirow{2}*{miniImageNet} & ICRL-Net & 65.55 $\pm$ 0.79	& 81.87 $\pm$ 0.51\\
      ~ & ICRL-Net+T & 67.33 $\pm$ 0.63 &82.65 $\pm$ 0.53 \\

       \multirow{2}*{tieredImageNet} & ICRL-Net & 70.56 $\pm$ 0.91	& 85.62 $\pm$ 0.64 \\
      ~ & ICRL-Net+T & 72.20 $\pm$ 0.80	& 86.05 $\pm$ 0.58 \\
     \hline
  \end{tabular}
 \label{tab:transductive}
\end{table}

\subsection{Transductive FSL}
We further extend the proposed method for transductive FSL by integrating our AIRN with the transductive inference algorithm developed by CAN~\cite{hou2019cross}. Specifically, we firstly utilize the proposed AIRN to obtain the class features and utilize the initial class features to predict the labels of the unlabeled query samples. Then we select several pseudo-labeled unlabeled query samples according to the label confidence criterion proposed by CAN~\cite{hou2019cross}. Finally, the selected pseudo-labeled unlabeled query samples are used together with the initial class features to generate more representative class features. From Table~\ref{tab:transductive}, we can see that performance of the proposed ICRL-Net under the transductive setting (ICRL-Net+T) are further improved.

\section{Conclusion}\label{sec:Conclusion}
In this paper, we present a novel framework for metric-based FSL by considering the information diversity of instances. The designed instance-adaptive class representation learning network (ICRL-Net) can automatically learn the relative significance of different support instances that belong to the same class. We conducted extensive experiments on four few-shot visual recognition benchmarks, and from the results, we mainly conclude that: 1) exploiting the significance of different instances is critical in obtaining the class-level representation, and the proposed ICRL-Net can adaptively and effectively evaluate the relative significance of different instances; 2) the proposed method outperforms the state-of-the-art FSL counterparts by a large margin, and all the different modules in our ICRL-Net are essential to achieve satisfactory performance. One disadvantage of our method may be that the importance is assessed at the instance level, while the relative importance of the different regions in a visual instance is not considered. How to revaluing the significance of different regions insider one instance remains a potential future work.

\section*{Acknowledgment}
The authors would like to thank the handling associate editor and all the anonymous reviewers for their constructive comments. This research was supported in part by the National Key Research and Development Program of China under No. 2021YFC3300200,
%Major Science and Technology Innovation 2030 ``New Generation Artificial Intelligence'' key project (No. 2021ZD0111700),
the Special Fund of Hubei Luojia Laboratory under Grant 220100014, and the National Natural Science Foundation of China (Grant No. 62002090 and 62141112).

%\bibliographystyle{IEEEtran}
%\balance
%\bibliography{IEEEabrv,TNNLS-2021-P-18223}

\begin{thebibliography}{100}

\bibitem{he2016deep}
Kaiming He, Xiangyu Zhang, Shaoqing Ren, and Jian Sun.
\newblock Deep residual learning for image recognition.
\newblock In {\em Proceedings of the IEEE Conference on Computer Vision and
  Pattern Recognition (CVPR)}, pages 770--778, 2016.
  
 \bibitem{huang2017densely}
Gao Huang, Zhuang Liu, Laurens Van Der~Maaten, and Kilian~Q Weinberger.
\newblock Densely connected convolutional networks.
\newblock In {\em Proceedings of the IEEE Conference on Computer Vision and
  Pattern Recognition (CVPR)}, pages 4700--4708, 2017.
  
 \bibitem{hu2018squeeze}
Jie Hu, Li~Shen, and Gang Sun.
\newblock Squeeze-and-excitation networks.
\newblock In {\em Proceedings of the IEEE Conference on Computer Vision and
  Pattern Recognition (CVPR)}, pages 7132--7141, 2018.
  
 \bibitem{redmon2016you}
Joseph Redmon, Santosh Divvala, Ross Girshick, and Ali Farhadi.
\newblock You only look once: Unified, real-time object detection.
\newblock In {\em Proceedings of the IEEE Conference on Computer Vision and
  Pattern Recognition (CVPR)}, pages 779--788, 2016.
  
 \bibitem{ren2015faster}
Shaoqing Ren, Kaiming He, Ross Girshick, and Jian Sun.
\newblock Faster r-cnn: Towards real-time object detection with region proposal
  networks.
\newblock In {\em Advances in Neural Information Processing Systems (NeurIPS)},
  pages 91--99, 2015.
  
\bibitem{he2017mask}
Kaiming He, Georgia Gkioxari, Piotr Doll{\'a}r, and Ross Girshick.
\newblock Mask r-cnn.
\newblock In {\em Proceedings of the IEEE International Conference on Computer
  Vision (ICCV)}, pages 2961--2969, 2017.
 
\bibitem{fe2003bayesian}
Li~Fe-Fei et~al.
\newblock A bayesian approach to unsupervised one-shot learning of object
  categories.
\newblock In {\em Proceedings Ninth IEEE International Conference on Computer
  Vision}, pages 1134--1141. IEEE, 2003.

\bibitem{fei2006one}
Li~Fei-Fei, Rob Fergus, and Pietro Perona.
\newblock One-shot learning of object categories.
\newblock {\em IEEE Transactions on Pattern Analysis and Machine Intelligence
  (TPAMI)}, 28(4):594--611, 2006.
 
\bibitem{gidaris2018dynamic}
Spyros Gidaris and Nikos Komodakis.
\newblock Dynamic few-shot visual learning without forgetting.
\newblock In {\em Proceedings of the IEEE Conference on Computer Vision and
  Pattern Recognition (CVPR)}, pages 4367--4375, 2018.
  
\bibitem{8948295}
Hong-Gyu Jung and Seong-Whan Lee.
\newblock Few-shot learning with geometric constraints.
\newblock {\em IEEE Transactions on Neural Networks and Learning Systems},
  31(11):4660--4672, 2020.  
  
\bibitem{9159937}
Nan Lai, Meina Kan, Chunrui Han, Xingguang Song, and Shiguang Shan.
\newblock Learning to learn adaptive classifier–predictor for few-shot
  learning.
\newblock {\em IEEE Transactions on Neural Networks and Learning Systems},
  32(8):3458--3470, 2021.
  
 \bibitem{lee2020self}
Hankook Lee, Sung~Ju Hwang, and Jinwoo Shin.
\newblock Self-supervised label augmentation via input transformations.
\newblock In {\em International Conference on Machine Learning (ICML)}, pages
  5714--5724, 2020.  

\bibitem{li2021multi}
Hainan Li, Renshuai Tao, Jun Li, Haotong Qin, Yifu Ding, Shuo Wang, and
  Xianglong Liu.
\newblock Multi-pretext attention network for few-shot learning with
  self-supervision.
\newblock In {\em 2021 IEEE International Conference on Multimedia and Expo
  (ICME)}, pages 1--6. IEEE, 2021.

\bibitem{9072304}
Jiang Lu, Sheng Jin, Jian Liang, and Changshui Zhang.
\newblock Robust few-shot learning for user-provided data.
\newblock {\em IEEE Transactions on Neural Networks and Learning Systems},
  32(4):1433--1447, 2021.

\bibitem{Luo_2021_WACV}
Qinxuan Luo, Lingfeng Wang, Jingguo Lv, Shiming Xiang, and Chunhong Pan.
\newblock Few-shot learning via feature hallucination with variational
  inference.
\newblock In {\em Proceedings of the IEEE/CVF Winter Conference on Applications
  of Computer Vision (WACV)}, pages 3963--3972, January 2021.

\bibitem{zhang2020uncertainty}
Zhizheng Zhang, Cuiling Lan, Wenjun Zeng, Zhibo Chen, and Shih-Fu Chang.
\newblock Uncertainty-aware few-shot image classification.
\newblock {\em IJCAI}, 2021.

\bibitem{pmlr-v119-ziko20a}
Imtiaz Ziko, Jose Dolz, Eric Granger, and Ismail~Ben Ayed.
\newblock {L}aplacian regularized few-shot learning.
\newblock In {\em International Conference on Machine Learning (ICML)}, pages
  11660--11670, 2020.

\bibitem{lee2019meta}
Kwonjoon Lee, Subhransu Maji, Avinash Ravichandran, and Stefano Soatto.
\newblock Meta-learning with differentiable convex optimization.
\newblock In {\em Proceedings of the IEEE Conference on Computer Vision and
  Pattern Recognition (CVPR)}, pages 10657--10665, 2019.
  
 \bibitem{9174820}
Jinghang Li and Mengqi Hu.
\newblock Continuous model adaptation using online meta-learning for smart grid
  application.
\newblock {\em IEEE Transactions on Neural Networks and Learning Systems},
  32(8):3633--3642, 2021.
  
\bibitem{ravi2016optimization}
Sachin Ravi and Hugo Larochelle.
\newblock Optimization as a model for few-shot learning.
\newblock In {\em International Conference on Learning Representations (ICLR)},
  2017. 
  
\bibitem{vinyals2016matching}
Oriol Vinyals, Charles Blundell, Timothy Lillicrap, Daan Wierstra, et~al.
\newblock Matching networks for one shot learning.
\newblock In {\em Advances in Neural Information Processing Systems (NeurIPS)},
  pages 3630--3638, 2016.  
  
\bibitem{elsken2020meta}
Thomas Elsken, Benedikt Staffler, Jan~Hendrik Metzen, and Frank Hutter.
\newblock Meta-learning of neural architectures for few-shot learning.
\newblock In {\em Proceedings of the IEEE Conference on Computer Vision and
  Pattern Recognition (CVPR)}, pages 12365--12375, 2020.

\bibitem{finn2017model}
Chelsea Finn, Pieter Abbeel, and Sergey Levine.
\newblock Model-agnostic meta-learning for fast adaptation of deep networks.
\newblock In {\em International Conference on Machine Learning (ICML)}, page
  1126–1135, 2017.

\bibitem{rusu2018meta}
Andrei~A Rusu, Dushyant Rao, Jakub Sygnowski, Oriol Vinyals, Razvan Pascanu,
  Simon Osindero, and Raia Hadsell.
\newblock Meta-learning with latent embedding optimization.
\newblock In {\em International Conference on Learning Representations (ICLR)},
  2019.

\bibitem{luo2021few}
Qinxuan Luo, Lingfeng Wang, Jingguo Lv, Shiming Xiang, and Chunhong Pan.
\newblock Few-shot learning via feature hallucination with variational
  inference.
\newblock In {\em Proceedings of the IEEE/CVF Winter Conference on Applications
  of Computer Vision (WACV)}, pages 3963--3972, 2021.

\bibitem{2020Adaptive}
Christian Simon, Piotr Koniusz, Richard Nock, and Mehrtash Harandi.
\newblock Adaptive subspaces for few-shot learning.
\newblock In {\em Proceedings of the IEEE Conference on Computer Vision and
  Pattern Recognition (CVPR)}, 2020.

\bibitem{tian2020rethinking}
Yonglong Tian, Yue Wang, Dilip Krishnan, Joshua~B Tenenbaum, and Phillip Isola.
\newblock Rethinking few-shot image classification: a good embedding is all you
  need?
\newblock {\em arXiv preprint arXiv:2003.11539}, 2020.

\bibitem{2020A}
Yinbo Chen, Zhuang Liu, Huijuan Xu, Trevor Darrell, and Xiaolong Wang.
\newblock Meta-baseline: Exploring simple meta-learning for few-shot learning.
\newblock In {\em Proceedings of the IEEE International Conference on Computer
  Vision (ICCV)}, 2021.
  
\bibitem{snell2017prototypical}
Jake Snell, Kevin Swersky, and Richard Zemel.
\newblock Prototypical networks for few-shot learning.
\newblock In {\em Advances in Neural Information Processing Systems (NeurIPS)},
  pages 4077--4087, 2017.
  
 \bibitem{sung2018learning}
Flood Sung, Yongxin Yang, Li~Zhang, Tao Xiang, Philip~HS Torr, and Timothy~M
  Hospedales.
\newblock Learning to compare: Relation network for few-shot learning.
\newblock In {\em Proceedings of the IEEE Conference on Computer Vision and
  Pattern Recognition (CVPR)}, pages 1199--1208, 2018.  

\bibitem{ren2018meta}
Mengye Ren, Eleni Triantafillou, Sachin Ravi, Jake Snell, Kevin Swersky,
  Joshua~B Tenenbaum, Hugo Larochelle, and Richard~S Zemel.
\newblock Meta-learning for semi-supervised few-shot classification.
\newblock In {\em International Conference on Learning Representations (ICLR)},
  2018. 
  
\bibitem{bertinetto2018meta}
Luca Bertinetto, Joao~F Henriques, Philip~HS Torr, and Andrea Vedaldi.
\newblock Meta-learning with differentiable closed-form solvers.
\newblock In {\em International Conference on Learning Representations (ICLR)},
  2019.
  
\bibitem{oreshkin2018tadam}
Boris Oreshkin, Pau~Rodr{\'\i}guez L{\'o}pez, and Alexandre Lacoste.
\newblock Tadam: Task dependent adaptive metric for improved few-shot learning.
\newblock In {\em Advances in Neural Information Processing Systems (NeurIPS)},
  pages 721--731, 2018.  
  
\bibitem{gidaris2019generating}
Spyros Gidaris and Nikos Komodakis.
\newblock Generating classification weights with gnn denoising autoencoders for
  few-shot learning.
\newblock In {\em Proceedings of the IEEE Conference on Computer Vision and
  Pattern Recognition (CVPR)}, pages 21--30, 2019.  
  
 \bibitem{9056801}
Nikolaos Passalis, Alexandros Iosifidis, Moncef Gabbouj, and Anastasios Tefas.
\newblock Hypersphere-based weight imprinting for few-shot learning on embedded
  devices.
\newblock {\em IEEE Transactions on Neural Networks and Learning Systems},
  32(2):925--930, 2021.
  
 \bibitem{chen2021pareto}
Zhengyu Chen, Jixie Ge, Heshen Zhan, Siteng Huang, and Donglin Wang.
\newblock Pareto self-supervised training for few-shot learning.
\newblock In {\em Proceedings of the IEEE Conference on Computer Vision and
  Pattern Recognition (CVPR)}, pages 13663--13672, 2021.

\bibitem{guo2020attentive}
Yiluan Guo and Ngai-Man Cheung.
\newblock Attentive weights generation for few shot learning via information
  maximization.
\newblock In {\em Proceedings of the IEEE Conference on Computer Vision and
  Pattern Recognition (CVPR)}, pages 13499--13508, 2020.
  
\bibitem{li2020adversarial}
Kai Li, Yulun Zhang, Kunpeng Li, and Yun Fu.
\newblock Adversarial feature hallucination networks for few-shot learning.
\newblock In {\em Proceedings of the IEEE Conference on Computer Vision and
  Pattern Recognition (CVPR)}, pages 13470--13479, 2020. 
  
\bibitem{wang2021mtunet}
Bowen Wang, Liangzhi Li, Manisha Verma, Yuta Nakashima, Ryo Kawasaki, and
  Hajime Nagahara.
\newblock Mtunet: Few-shot image classification with visual explanations.
\newblock In {\em Proceedings of the IEEE Conference on Computer Vision and
  Pattern Recognition (CVPR)}, pages 2294--2298, 2021.  
  
\bibitem{li2017meta}
Zhenguo Li, Fengwei Zhou, Fei Chen, and Hang Li.
\newblock Meta-sgd: Learning to learn quickly for few-shot learning.
\newblock {\em arXiv preprint arXiv:1707.09835}, 2017.

\bibitem{mishra2017simple}
Nikhil Mishra, Mostafa Rohaninejad, Xi~Chen, and Pieter Abbeel.
\newblock A simple neural attentive meta-learner.
\newblock In {\em International Conference on Learning Representations (ICLR)},
  2018.

\bibitem{sun2019meta}
Qianru Sun, Yaoyao Liu, Tat-Seng Chua, and Bernt Schiele.
\newblock Meta-transfer learning for few-shot learning.
\newblock In {\em Proceedings of the IEEE Conference on Computer Vision and
  Pattern Recognition (CVPR)}, pages 403--412, 2019.
  
\bibitem{hariharan2017low}
Bharath Hariharan and Ross Girshick.
\newblock Low-shot visual recognition by shrinking and hallucinating features.
\newblock In {\em Proceedings of the IEEE International Conference on Computer
  Vision (ICCV)}, pages 3018--3027, 2017.

\bibitem{koch2015siamese}
Gregory Koch, Richard Zemel, and Ruslan Salakhutdinov.
\newblock Siamese neural networks for one-shot image recognition.
\newblock In {\em ICML deep learning workshop}, volume~2. Lille, 2015.

\bibitem{li2019revisiting}
Wenbin Li, Lei Wang, Jinglin Xu, Jing Huo, Yang Gao, and Jiebo Luo.
\newblock Revisiting local descriptor based image-to-class measure for few-shot
  learning.
\newblock In {\em Proceedings of the IEEE Conference on Computer Vision and
  Pattern Recognition (CVPR)}, pages 7260--7268, 2019.
  
\bibitem{2017Low}
Hang Qi, Matthew Brown, and David~G. Lowe.
\newblock Low-shot learning with imprinted weights.
\newblock In {\em Proceedings of the IEEE Conference on Computer Vision and
  Pattern Recognition (CVPR)}, 2017.
  
\bibitem{hou2019cross}
Ruibing Hou, Hong Chang, Bingpeng Ma, Shiguang Shan, and Xilin Chen.
\newblock Cross attention network for few-shot classification.
\newblock In {\em Advances in Neural Information Processing Systems (NeurIPS)},
  pages 4005--4016, 2019.

\bibitem{ye2020few}
Han-Jia Ye, Hexiang Hu, De-Chuan Zhan, and Fei Sha.
\newblock Few-shot learning via embedding adaptation with set-to-set functions.
\newblock In {\em Proceedings of the IEEE Conference on Computer Vision and
  Pattern Recognition (CVPR)}, pages 8808--8817, 2020.  
  
\bibitem{kye2020meta}
Seong~Min Kye, Hae~Beom Lee, Hoirin Kim, and Sung~Ju Hwang.
\newblock Meta-learned confidence for few-shot learning.
\newblock {\em arXiv preprint arXiv:2002.12017}, 2020.

\bibitem{zhao2021remp}
Yang Zhao, Chunyuan Li, Ping Yu, and Changyou Chen.
\newblock Remp: Rectified metric propagation for few-shot learning.
\newblock In {\em Proceedings of the IEEE Conference on Computer Vision and
  Pattern Recognition (CVPR) Workshops}, pages 2581--2590, 2021.

\bibitem{wang2020instance}
Yikai Wang, Chengming Xu, Chen Liu, Li~Zhang, and Yanwei Fu.
\newblock Instance credibility inference for few-shot learning.
\newblock In {\em Proceedings of the IEEE Conference on Computer Vision and
  Pattern Recognition (CVPR)}, pages 12836--12845, 2020.

\bibitem{wang2021trust}
Yikai Wang, Li~Zhang, Yuan Yao, and Yanwei Fu.
\newblock How to trust unlabeled data instance credibility inference for
  few-shot learning.
\newblock {\em IEEE Transactions on Pattern Analysis and Machine Intelligence},
  2021.
  
\bibitem{Wang_2018_CVPR}
Xiaolong Wang, Ross Girshick, Abhinav Gupta, and Kaiming He.
\newblock Non-local neural networks.
\newblock In {\em Proceedings of the IEEE Conference on Computer Vision and
  Pattern Recognition (CVPR)}, June 2018.
  
\bibitem{woo2018cbam}
Sanghyun Woo, Jongchan Park, Joon-Young Lee, and In~So Kweon.
\newblock Cbam: Convolutional block attention module.
\newblock In {\em Proceedings of the European Conference on Computer Vision
  (ECCV)}, pages 3--19, 2018.

\bibitem{velivckovic2017graph}
Petar Veli{\v{c}}kovi{\'c}, Guillem Cucurull, Arantxa Casanova, Adriana Romero,
  Pietro Lio, and Yoshua Bengio.
\newblock Graph attention networks.
\newblock {\em arXiv preprint arXiv:1710.10903}, 2017.

\bibitem{han2021transformer}
Kai Han, An~Xiao, Enhua Wu, Jianyuan Guo, Chunjing Xu, and Yunhe Wang.
\newblock Transformer in transformer.
\newblock {\em arXiv preprint arXiv:2103.00112}, 2021.

\bibitem{vaswani2017attention}
Ashish Vaswani, Noam Shazeer, Niki Parmar, Jakob Uszkoreit, Llion Jones,
  Aidan~N Gomez, {\L}ukasz Kaiser, and Illia Polosukhin.
\newblock Attention is all you need.
\newblock In {\em Advances in Neural Information Processing Systems (NeurIPS)},
  pages 5998--6008, 2017.

\bibitem{Wang_2017_CVPR}
Peng Wang, Lingqiao Liu, Chunhua Shen, Zi~Huang, Anton van~den Hengel, and Heng
  Tao~Shen.
\newblock Multi-attention network for one shot learning.
\newblock In {\em Proceedings of the IEEE Conference on Computer Vision and
  Pattern Recognition (CVPR)}, July 2017.
  
\bibitem{xu2021attentional}
Weijian Xu, Huaijin Wang, Zhuowen Tu, et~al.
\newblock Attentional constellation nets for few-shot learning.
\newblock In {\em International Conference on Learning Representations (ICLR)},
  2021.

\bibitem{ionescu2015matrix}
Catalin Ionescu, Orestis Vantzos, and Cristian Sminchisescu.
\newblock Matrix backpropagation for deep networks with structured layers.
\newblock In {\em Proceedings of the IEEE International Conference on Computer
  Vision (ICCV)}, pages 2965--2973, 2015.

\bibitem{kim2016hadamard}
Jin-Hwa Kim, Kyoung-Woon On, Woosang Lim, Jeonghee Kim, Jung-Woo Ha, and
  Byoung-Tak Zhang.
\newblock Hadamard product for low-rank bilinear pooling.
\newblock {\em arXiv preprint arXiv:1610.04325}, 2016.

\bibitem{lin2015bilinear}
Tsung-Yu Lin, Aruni RoyChowdhury, and Subhransu Maji.
\newblock Bilinear cnn models for fine-grained visual recognition.
\newblock In {\em Proceedings of the IEEE International Conference on Computer
  Vision (ICCV)}, pages 1449--1457, 2015.
  
\bibitem{lin2013network}
Min Lin, Qiang Chen, and Shuicheng Yan.
\newblock Network in network.
\newblock {\em arXiv preprint arXiv:1312.4400}, 2013.

\bibitem{zhang2019power}
Hongguang Zhang and Piotr Koniusz.
\newblock Power normalizing second-order similarity network for few-shot
  learning.
\newblock In {\em 2019 IEEE winter conference on applications of computer
  vision (WACV)}, pages 1185--1193. IEEE, 2019.
  
 \bibitem{zhang2019few}
Hongguang Zhang, Jing Zhang, and Piotr Koniusz.
\newblock Few-shot learning via saliency-guided hallucination of samples.
\newblock In {\em Proceedings of the IEEE Conference on Computer Vision and
  Pattern Recognition (CVPR)}, pages 2770--2779, 2019.

\bibitem{zhang2020few}
Hongguang Zhang, Philip~HS Torr, and Piotr Koniusz.
\newblock Few-shot learning with multi-scale self-supervision.
\newblock {\em arXiv preprint arXiv:2001.01600}, 2020.

\bibitem{wertheimer2019few}
Davis Wertheimer and Bharath Hariharan.
\newblock Few-shot learning with localization in realistic settings.
\newblock In {\em Proceedings of the IEEE Conference on Computer Vision and
  Pattern Recognition (CVPR)}, pages 6558--6567, 2019.
  
\bibitem{koniusz2021power}
Piotr Koniusz and Hongguang Zhang.
\newblock Power normalizations in fine-grained image, few-shot image and graph
  classification.
\newblock {\em IEEE Transactions on Pattern Analysis and Machine Intelligence},
  2021.


\bibitem{goldblum2020unraveling}
Micah Goldblum, Steven Reich, Liam Fowl, Renkun Ni, Valeriia Cherepanova, and
  Tom Goldstein.
\newblock Unraveling meta-learning: Understanding feature representations for
  few-shot tasks.
\newblock In {\em International Conference on Machine Learning (ICML)}, pages
  3607--3616. PMLR, 2020.
  
 \bibitem{russakovsky2015imagenet}
Olga Russakovsky, Jia Deng, Hao Su, Jonathan Krause, Sanjeev Satheesh, Sean Ma,
  Zhiheng Huang, Andrej Karpathy, Aditya Khosla, Michael Bernstein, et~al.
\newblock Imagenet large scale visual recognition challenge.
\newblock {\em International journal of computer vision}, 115(3):211--252,
  2015.
  
 \bibitem{krizhevsky2010cifar}
Alex Krizhevsky, Vinod Nair, and Geoffrey Hinton.
\newblock Cifar-100 (canadian institute for advanced research).
\newblock {\em URL http://www. cs. toronto. edu/kriz/cifar. html}. 

\bibitem{8935497}
Debasmit Das and C.~S.~George Lee.
\newblock A two-stage approach to few-shot learning for image recognition.
\newblock {\em IEEE Transactions on Image Processing}, 29:3336--3350, 2020.

\bibitem{zhang2020deepemd}
Chi Zhang, Yujun Cai, Guosheng Lin, and Chunhua Shen.
\newblock Deepemd: Few-shot image classification with differentiable earth
  mover's distance and structured classifiers.
\newblock In {\em Proceedings of the IEEE Conference on Computer Vision and
  Pattern Recognition (CVPR)}, pages 12203--12213, 2020.
  
 \bibitem{kim2020model}
Jaekyeom Kim, Hyoungseok Kim, and Gunhee Kim.
\newblock Model-agnostic boundary-adversarial sampling for test-time
  generalization in few-shot learning.
\newblock In {\em Proceedings of the European Conference on Computer Vision
  (ECCV)}, 2020.

\bibitem{chen2019closer}
Wei-Yu Chen, Yen-Cheng Liu, Zsolt Kira, Yu-Chiang~Frank Wang, and Jia-Bin
  Huang.
\newblock A closer look at few-shot classification.
\newblock {\em arXiv preprint arXiv:1904.04232}, 2019.

\end{thebibliography}

\begin{IEEEbiography}[{\includegraphics[width=1in,height=1.25in,clip,keepaspectratio]{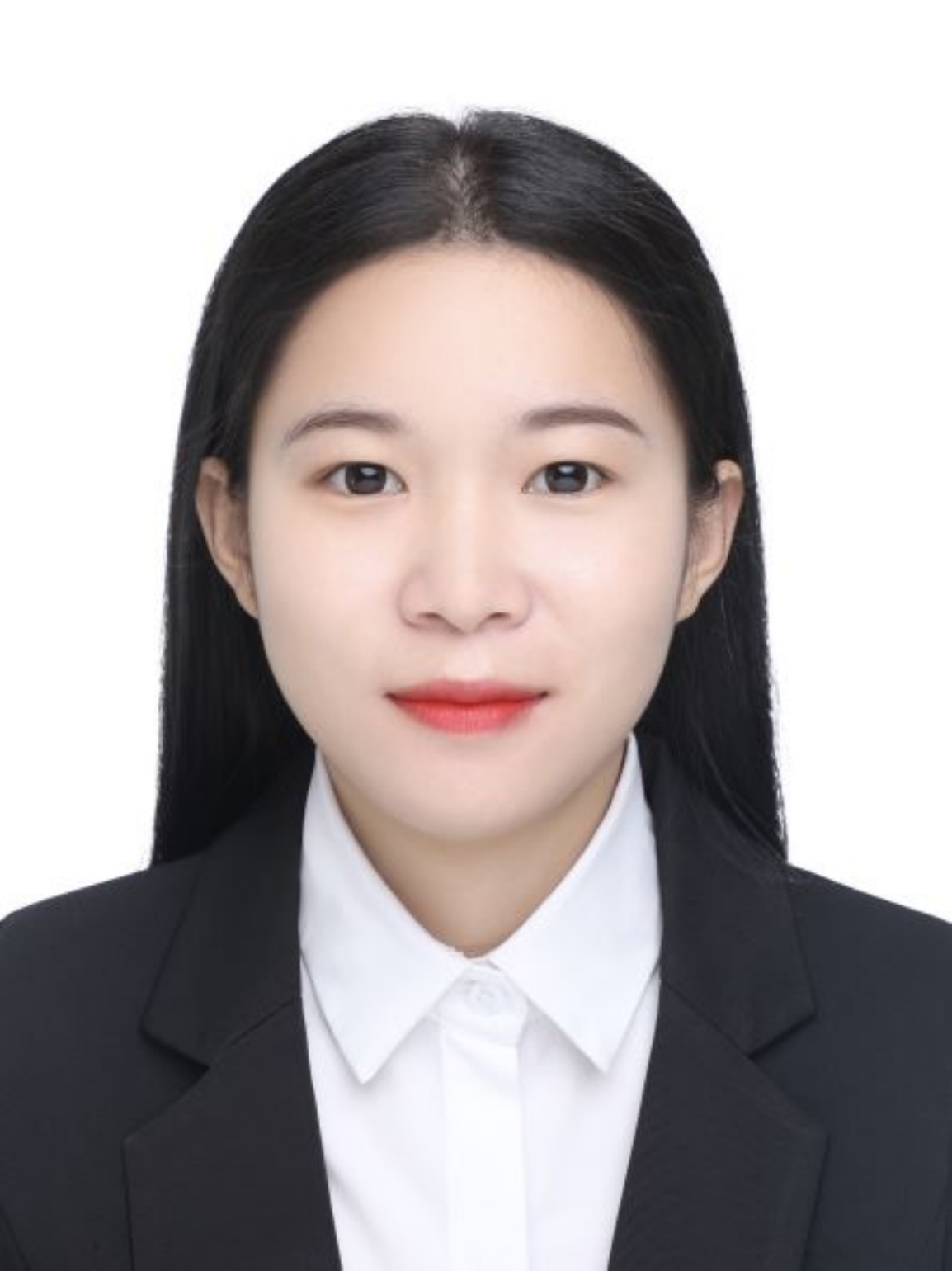}}]{Mengya Han}
received the B.S. degree in Computer Science from the Bengbu University, Bengbu, China, and the M.S degree in the School of Computer Science and Information Engineering, Hefei University of Technology, Hefei, China. She is currently a PhD student in the School of Computer Science, Wuhan University, China. Her research interests are primarily in computer vision and machine learning.
\end{IEEEbiography}

\begin{IEEEbiography}[{\includegraphics[width=1in,height=1.25in,clip,keepaspectratio]{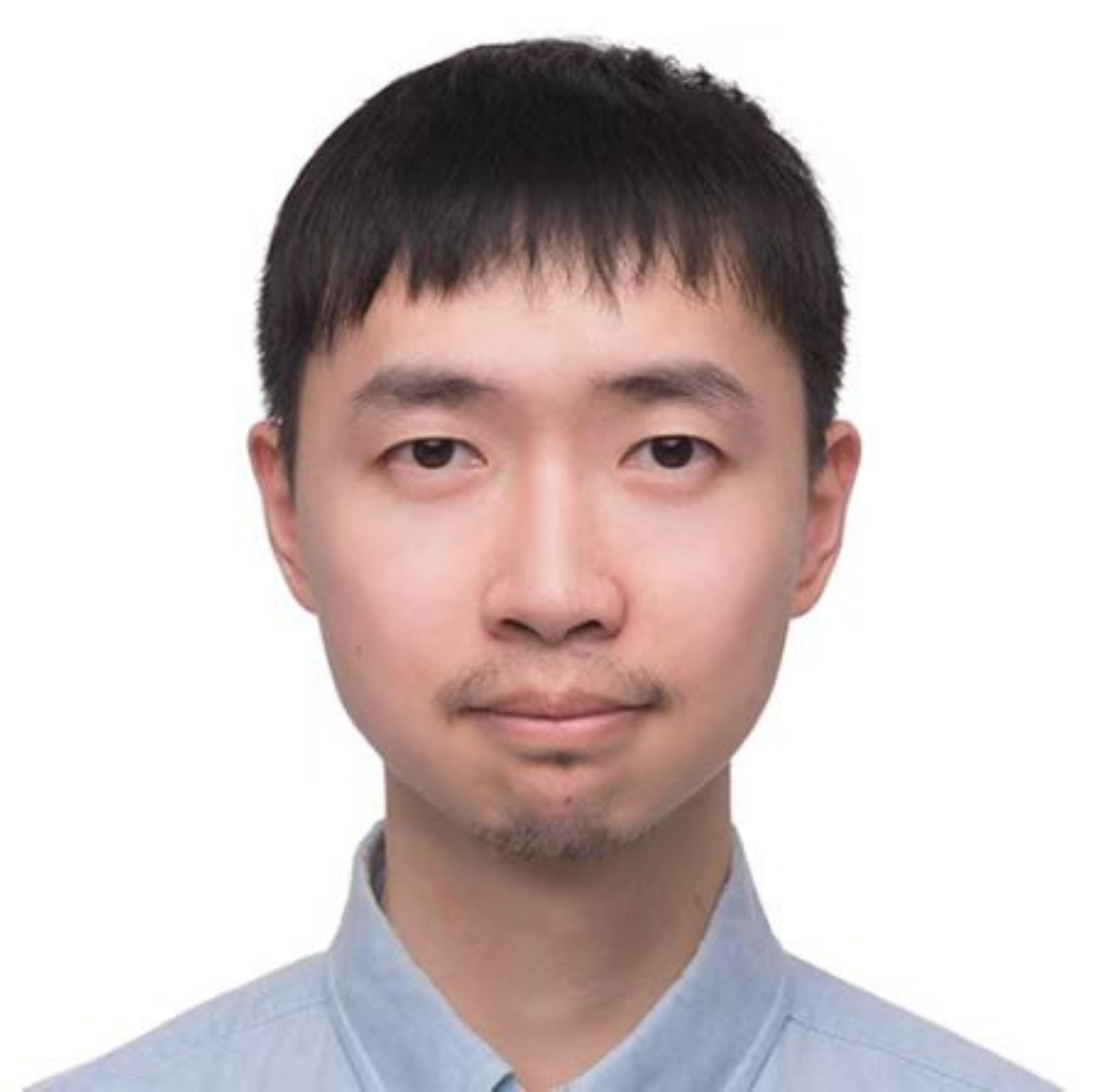}}]{Yibing Zhan} received a B.E. and a Ph.D. from the University of Science and Technology of China in 2012 and 2018, respectively. From 2018 to 2020, he was an Associate Researcher in the Computer and Software School at the Hangzhou Dianzi University. He is currently an algorithm scientist at the JD Explore Academy. His research interest is in graphical models and multimodal learning, including cross-modal retrieval, scene graph generation, and graph neural networks. He has publications on various top conferences and journals, such as CVPR, ACM MM, AAAI, IJCV, and IEEE TMM.
\end{IEEEbiography}

\begin{IEEEbiography}[{\includegraphics[width=1in,height=1.25in,clip,keepaspectratio]{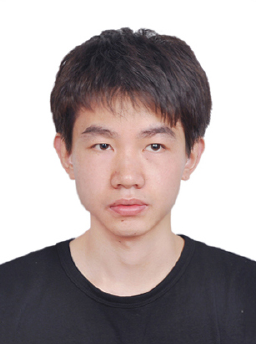}}]{Yong Luo}
received the B.E. degree in Computer Science from the Northwestern Polytechnical University, Xi'an, China, and the D.Sc. degree in the School of Electronics Engineering and Computer Science, Peking University, Beijing, China. He is currently a Professor with the School of Computer Science, Wuhan University, China. His research interests are primarily on machine learning and data mining with applications to visual information understanding and analysis. He has authored or co-authored over 60 papers in top journals and prestigious conferences including IEEE T-PAMI, IEEE T-NNLS, IEEE T-IP, IEEE T-KDE, IEEE T-MM, ICCV, WWW, IJCAI and AAAI. He is serving on editorial board for IEEE T-MM. He received the IEEE Globecom 2016 Best Paper Award, and was nominated as the IJCAI 2017 Distinguished Best Paper Award. He is also a co-recipient of the IEEE ICME 2019 and IEEE VCIP 2019 Best Paper Awards.
\end{IEEEbiography}

\begin{IEEEbiography}[{\includegraphics[width=1in,height=1.25in,clip,keepaspectratio]{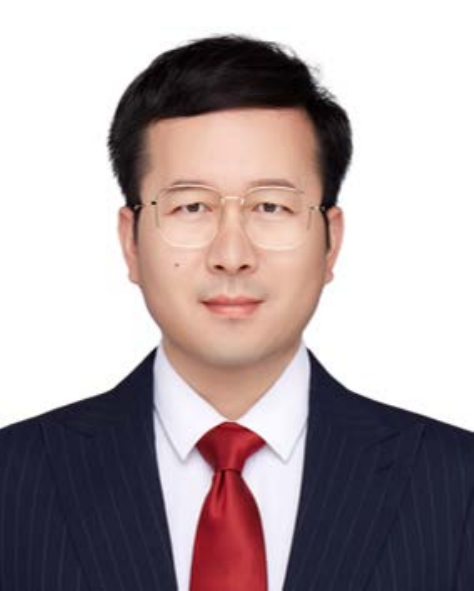}}]{Bo Du} received the PhD degree from the State Key Laboratory of Information Engineering in Surveying, Mapping and Remote Sensing, Wuhan University, Wuhan, China, in 2010. He is currently a professor with the School of Computer, Wuhan University. He has published more than 100 scientific papers, such as the IEEE Transactions on Geoscience and Remote Sensing, IEEE Transactions on Neural Networks and Learning Systems, IEEE Transactions on Image Processing, IEEE Transactions on Cybernetics, AAAI, and IJCAI. His research interests include pattern recognition, hyperspectral image processing, and signal processing.
\end{IEEEbiography}

\begin{IEEEbiography}[{\includegraphics[width=1in,height=1.25in,clip,keepaspectratio]{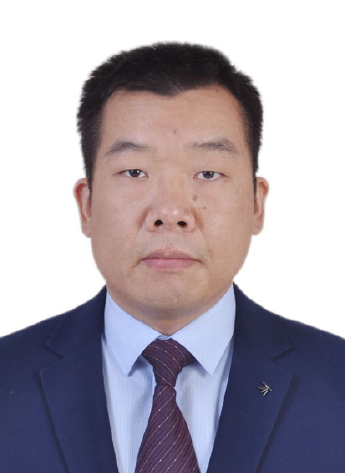}}]{Han Hu}
received the B.E. and Ph.D. degrees from the University of Science and Technology of China, China, in 2007 and 2012, respectively. He is currently a Professor with the School of Information and Electronics, Beijing Institute of Technology, China. His research interests include multimedia networking, edge intelligence, and space-air-ground integrated network. He received several academic awards, including the Best Paper Award of the IEEE TCSVT 2019, the Best Paper Award of the IEEE Multimedia Magazine 2015, and the Best Paper Award of the IEEE Globecom 2013. He has served as an Associate Editor of IEEE TMM and Ad Hoc Networks, and a TPC member of Infocom, ACM MM, AAAI, and IJCAI.
\end{IEEEbiography}

\begin{IEEEbiography}[{\includegraphics[width=1in,height=1.25in,clip,keepaspectratio]{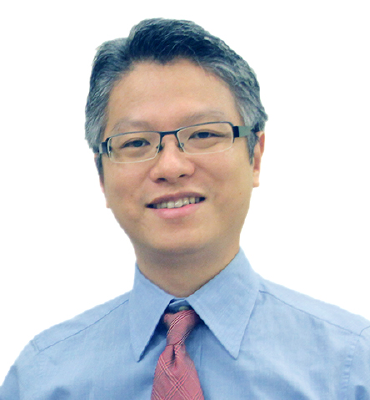}}]{Yonggang Wen}
(M'08-S'14-F'20) is the Professor of Computer Science and Engineering at Nanyang Technological University (NTU), Singapore. He has also served as the Associate Dean (Research) at College of Engineering at NTU Singapore since 2018. He served as the Acting Director of Nanyang Technopreneurship Centre (NTC) at NTU from 2017 to 2019, and the Assistant Chair (Innovation) of School of Computer Science and Engineering (SCSE) at NTU from 2016 to 2018. He received his PhD degree in Electrical Engineering and Computer Science (minor in Western Literature) from Massachusetts Institute of Technology (MIT), Cambridge, USA, in 2008. He is a co-recipient of multiple journal best papers awards, including IEEE Transactions on Circuits and Systems for Video Technology (2019), IEEE Multimedia (2015), and and several best paper awards from international conferences, including 2020 IEEE VCIP, 2016 IEEE Globecom, 2016 IEEE Infocom MuSIC Workshop, 2015 EAI/ICST Chinacom, 2014 IEEE WCSP, 2013 IEEE Globecom and 2012 IEEE EUC. He received 2016 IEEE ComSoc MMTC Distinguished Leadership Award. His research interests include cloud computing, green data center, distributed machine learning, blockchain, big data analytics, multimedia network and mobile computing. He is a Fellow of IEEE.
\end{IEEEbiography}

\begin{IEEEbiography}[{\includegraphics[width=1in,height=1.25in,clip,keepaspectratio]{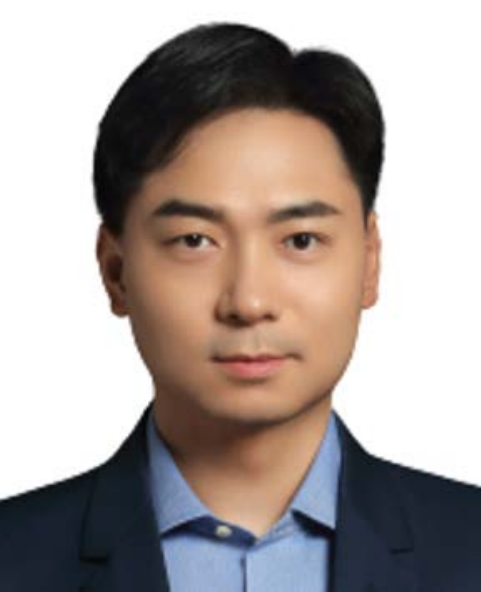}}]{Dacheng Tao}
(F'15) is currently an Advisor and a Chief Scientist of the Digital Science Institute, Faculty of Engineering, University of Sydney, Darlington, NSW, Australia. He is also the Director of the JD Explore Academy and a Vice President of JD.com. He mainly applies statistics and mathematics to artificial intelligence and data science, and his research is detailed in one monograph and over 200 publications in prestigious journals and proceedings at leading conferences. Dr. Tao received the 2015 Australian ScopusEureka Prize, the 2018 IEEE ICDM Research Contributions Award, and the 2021 IEEE Computer Society McCluskey Technical Achievement Award. He is a Fellow of the Australian Academy of Science, AAAS, and ACM.
\end{IEEEbiography}

\end{document}